\documentclass{article}

\PassOptionsToPackage{numbers, sort&compress}{natbib}
\usepackage[main, final]{neurips_2026}
\newcommand{\MethodName}{G2PO}
\usepackage{amsmath}
\usepackage{amssymb}
\usepackage{amsthm}
\usepackage{multirow}
\usepackage[table]{xcolor}
\usepackage{graphicx}
\usepackage{wrapfig}
\usepackage{enumitem}
\usepackage{algorithm}
\usepackage{algorithmic}
\usepackage{tcolorbox}
\usepackage{subcaption}

\definecolor{midnightgreen}{rgb}{0.0, 0.29, 0.33}
\definecolor{deepgreen}{HTML}{055c29}
\definecolor{deeppurple}{HTML}{7030a0}
\definecolor{deepblue}{HTML}{171d91}
\definecolor{brown}{HTML}{843c0c}
\definecolor{shadered}{HTML}{ffe5e5}
\definecolor{shadegreen}{HTML}{e5f7ed}
\definecolor{msftBlack}{RGB}{0,0,0}
\definecolor{lightred}{RGB}{255,163,163}
\definecolor{deepred}{RGB}{153,0,0}
\definecolor{barblue}{RGB}{90,120,180}
\definecolor{barorange}{RGB}{225,124,5}
\definecolor{softblue}{RGB}{30, 90, 160}
\usepackage{hyperref}       
\hypersetup{
  colorlinks=true,
  linkcolor=deepred,
  citecolor=softblue,
  urlcolor=magenta
}

\usepackage[utf8]{inputenc} 
\usepackage[T1]{fontenc}    
\usepackage{hyperref}       
\usepackage{url}            
\usepackage{booktabs}       
\usepackage{amsfonts}       
\usepackage{nicefrac}       
\usepackage{microtype}      
\usepackage{xcolor}         

\title{Group-Graph Policy Optimization for Long-Horizon Agentic Reinforcement Learning}

\author{%
  Yunan Wang$^1$, Minghui Song$^2$, Zihan Zhang$^2$, Shaohan Huang$^2$, \\ 
  \textbf{Haizhen Huang$^2$, Furu Wei$^2$, Weiwei Deng$^2$, Feng Sun$^2$, Qi Zhang$^2$} \\
  $^1$Peking University \qquad $^2$Microsoft Corporation \\
  \texttt{yunanwang25@stu.pku.edu.cn}
}

\begin{document}

\maketitle

\begin{abstract}
Group-based Reinforcement Learning (RL) has significantly enhanced Large Language Models (LLMs) in agentic scenarios. To achieve finer-grained policy updates, recent agentic RL frameworks have shifted from trajectory-level to step-level training. However, long-horizon agentic RL suffers from severe reward sparsity and delay, as feedback is often deferred for dozens of interaction steps. While existing step-level frameworks refine training granularity, their credit assignment remains coarse-grained and still treats agent exploration as isolated, linear trajectories. This oversimplified perspective ignores the inherent graph structure of state transitions, leading to high-variance state-value estimation and myopic, localized credit assignment. To overcome these critical bottlenecks, we propose Group-Graph Policy Optimization (\MethodName{}), a novel group-based RL algorithm tailored for multi-turn agentic tasks. \MethodName{} explicitly transforms linear interaction trajectories into a global state-transition graph. By aggregating identical observations across different trajectories, we introduce group-aggregation state-value estimation that reduces sampling variance and trajectory-dependent bias. Furthermore, we redefine agent actions as transitions between state nodes and propose an edge-centric advantage estimation strategy. By globally standardizing Temporal Difference (TD) errors across the entire graph, \MethodName{} explicitly identifies and prioritizes critical transitions that drive absolute task progress. Extensive experiments on representative long-horizon benchmarks—WebShop, ALFWorld, and AppWorld—demonstrate that \MethodName{} substantially outperforms state-of-the-art prompt-based and RL baselines, achieving remarkable success rate improvements of up to 22.2\% over GRPO.  Code available at \url{https://github.com/Nala-YN/G2PO}.
\end{abstract}

\section{Introduction}

The field of Large Language Models (LLMs) is witnessing a transformative leap from Chatbot to Copilot, and ultimately to \textit{Agent} \cite{luo2025large,huang2024understanding,ferrag2025llm}. By interacting with real-world external environments (e.g., e-commerce websites \cite{yao2022webshop}, embodied physical simulations \cite{shridhar2020alfworld}, and complex software ecosystems \cite{trivedi2024appworld}), LLM agents are increasingly demonstrating the long-horizon decision-making capabilities to solve complex, real-world problems. 

\begin{figure}[!t]
\centering
\includegraphics[width=1.0\columnwidth]{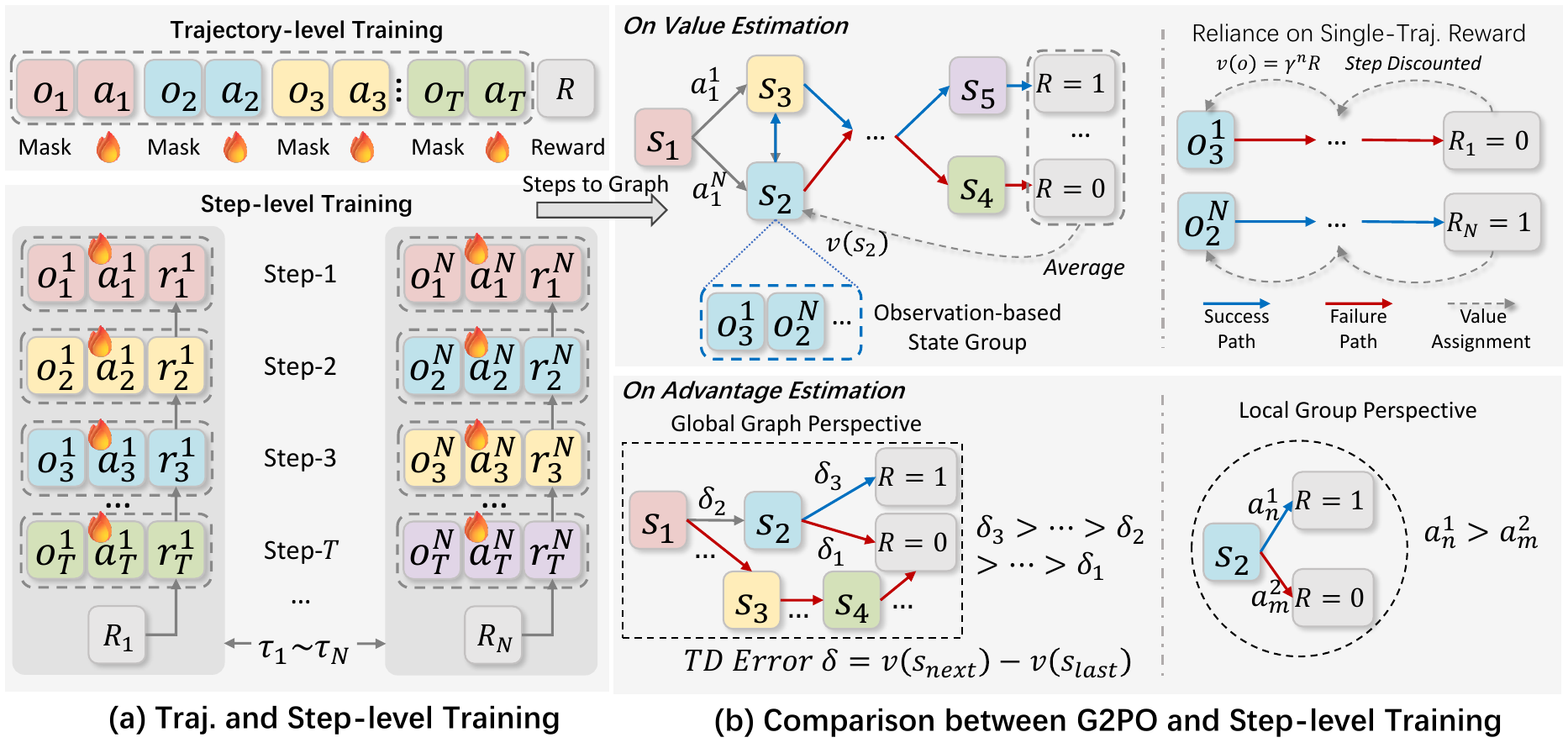}
\caption{(a) compares the training sample units between trajectory-level and step-level training. (b) contrasts \MethodName{} (left) with existing step-level RL methods (right): while current approaches treat agent exploration as a collection of linear trajectories for credit assignment, \MethodName{} tackles credit assignment from a global graph perspective.}
\label{fig:intro}
\vspace{-0.5cm}
\end{figure}

To enhance LLM reasoning, Reinforcement Learning (RL) has become a standard post-training pipeline for frontier models like GPT-5.2 \cite{gpt-5.2}, Gemini 3 Pro \cite{gemini-3-pro}, and DeepSeek V3.2 \cite{liu2025deepseek}. Through algorithms such as PPO \cite{schulman2017proximal}, GRPO \cite{shao2024deepseekmath}, SAPO \cite{gao2025soft}, and DAPO \cite{yu2025dapo}, RL has drastically improved model reasoning on math and coding tasks \cite{guo2025deepseek,cao2026qwen3,ren2025deepseek}. However, unlike single-turn reasoning, agentic RL typically requires the agent to interact with the environment for dozens of turns for a complete trajectory. Existing trajectory-level agentic RL frameworks, such as RAGEN \cite{wang2025ragen} and Search-R1 \cite{jinsearch}, concatenate all environmental observations and policy actions into a single training sample. As the number of interaction turns increases, this trajectory-level training strategy leads to an explosive growth in context length and extreme memory consumption. To address this, recent frameworks \cite{wang2025reinforcement,yu2025memagent,chen2025context,luo2025agent} have shifted to step-level training. As shown in Figure \ref{fig:intro}(a), by treating each step's observation and action as an individual training sample, they resolve the context explosion problem and enable lightweight agentic RL. To further provide step-level advantage, GiGPO \cite{fenggroup} adapts GRPO by calculating the relative advantage of actions within a group that shares the same environment observation. 

Despite the refinement of training granularity, existing step-level RL methods still adopt coarse-grained credit assignment. These methods treat agent exploration as a collection of isolated, linear trajectories. This perspective fails to capture the true dynamics of multi-turn interactions, where different trajectories often encounter the same environmental states (e.g., the same webpage or physical configuration). These shared states naturally act as nodes to connect trajectories, with actions serving as edges that represent state transitions. Consequently, the underlying structure of a long-horizon task is inherently a state-transition graph. This structural oversight leads to two critical limitations in step-level training, as illustrated in Figure \ref{fig:intro}(b).

\textbf{Limitation (i): High variance in state-value estimation.} Due to the stochastic nature of sampling, different trajectories starting from the same state node often traverse distinct paths on the graph and terminate at different end-states. Evaluating a state's value based on only one of these possible outcomes is inherently unreliable. For example, a trajectory starting from a promising state might still fail due to subsequent errors, while one starting from a poor state might succeed through later corrections. Evaluating a state based on a single sample makes the estimation highly susceptible to chance and leads to high variance, as shown in Figure \ref{fig:var}.

\textbf{Limitation (ii): Limited comparative scope of local group paradigms.} Current methods (e.g., GiGPO) evaluate an action solely by comparing it to other available actions at the same local state. While this captures local preferences, it lacks the global perspective to resolve credit assignment across the entire graph horizon. Consequently, local comparisons fail to account for the global importance of an action, making it impossible to distinguish between a minor improvement in a trivial state and a critical breakthrough action that significantly advances the overall task progress.

To resolve these issues, we propose Group-Graph Policy Optimization (\MethodName{}), a new group-based RL algorithm specifically designed for long-horizon agent tasks. We first explicitly reconstruct the sampled trajectories into a state-transition graph. Based on this graph, we introduce two novel designs: (i) \textit{Group-aggregation state-value estimation}: Unlike prior studies that rely on a single trajectory's reward, we aggregate intermediate steps with identical or similar observations across all trajectories into a \textit{state group}. The state value is then defined as the mean value of all states within this group. The key insight of this design is that the value of an environment state should be determined by its likelihood of leading to success, rather than a chance outcome from a single sample. This aligns with the underlying philosophy of GRPO, which similarly leverages the empirical mean of all sampled outcomes as a baseline. (ii) \textit{Edge-centric advantage estimation:} Within the state graph, agent actions function as directed edges representing state transitions. Having established robust value estimates for each state node, we can naturally evaluate an action based on the value increment between its source and destination nodes. In the context of sparse-reward agent tasks, this formulation aligns with the 1-step Temporal Difference (TD) error \cite{sutton2018reinforcement}. Crucially, rather than normalizing these advantages locally, we construct a global reference group comprising the TD errors of all transition edges across the entire graph. By globally standardizing these value increments, \MethodName{} explicitly prioritizes \textit{critical transitions} that drive the most substantial absolute progress toward the final goal.

We evaluate \MethodName{} on three representative long-horizon agent benchmarks: WebShop, ALFWorld, and AppWorld. Experiments on Qwen-2.5-1.5B/7B/14B demonstrate that \MethodName{} consistently outperforms existing prompt-based methods and RL training baselines. Notably, \MethodName{} achieves success rate improvements of up to 14.4\% and 22.2\% over GRPO on WebShop and ALFWorld, respectively. The AppWorld results further validate its effectiveness in tasks with complex environmental observations. Notably, \MethodName{} achieves these performance gains with almost no additional computational cost.

The main contributions of this paper are summarized as follows:
\vspace{-5pt}
\begin{itemize}[leftmargin=15pt, itemsep=0pt]
    \item We propose \MethodName, a group-based reinforcement learning algorithm specifically designed for long-horizon agent tasks. By transforming linear interaction trajectories into an environment state group graph, \MethodName{} enables fine-grained, step-wise training.
    \item We introduce group-aggregation state-value estimation to address the high variance in state-value estimation of existing group-based step-level RL methods.
    \item We redefine agent actions as transitions between nodes in the state group graph and design an edge-centric advantage estimation strategy that explicitly measures the absolute contribution of each step toward the global goal.
    \item We conduct extensive experiments across three demanding long-horizon agent benchmarks. \MethodName{} consistently outperforms state-of-the-art prompt-based and RL baselines with almost no computational overhead.
\end{itemize}

\section{Related work}
\textbf{LLM Agents.} LLMs have evolved from passive text generators to autonomous agents capable of engaging in multi-turn interactions with external environments \cite{wang2024executable,zhou2024language,hu2025agentgen,schmidgall2025agent}. Unlike single-turn QA tasks, agentic tasks (e.g., WebShop, AlfWorld) \cite{zhouwebarena,deng2023mind2web,xie2024osworld,mialon2023gaia} require the model to generate a sequence of actions over a long horizon to achieve a final goal. To solve agentic tasks, approaches such as ReAct \cite{yao2022react}, Toolformer \cite{schick2023toolformer}, and Reflexion \cite{shinn2023reflexion} demonstrate that interleaving reasoning traces with environment actions significantly enhances task performance. Agentic Reasoning \cite{wu2025agentic} enhances LLM reasoning capabilities by integrating autonomous agentic tools into an efficient and structured problem-solving process. MetaGPT \cite{hong2023metagpt} improves multi-agent collaboration by encoding human standard operating procedures into a structured assembly-line paradigm for complex task-solving.

\textbf{Reinforcement Learning for LLMs.} RL for LLMs has evolved from initial preference alignment using PPO-based RLHF \cite{ouyang2022training,bai2022training,christiano2017deep} and DPO \cite{rafailov2023direct,meng2024simpo,zhou2024wpo} to enhancing complex reasoning capabilities through emergent group-based RL algorithms. By estimating advantages from multiple samples and eliminating the need for a critic network, methods such as RLOO and GRPO have enabled memory-efficient, large-scale training. However, these approaches predominantly rely on sparse outcome-based rewards, which suffer from the credit assignment problem by failing to distinguish the contributions of individual reasoning steps. While process reward models \cite{caodreamprm,liprocess,cao2025more} offer step-level guidance, they often require prohibitive annotation costs and are prone to reward hacking. In contrast, \MethodName{} leverages a state transition graph to derive dense, multi-granularity advantages directly from sampling statistics, effectively solving the credit assignment problem without the computational overhead of a learned critic or the annotation costs of process supervision.

\textbf{Reinforcement Learning for LLM Agents.} RL has emerged as a cornerstone for empowering LLM agents to adapt and thrive in dynamic, open-ended environments \cite{webb2025brain,liu2025llm,zhang2025agentrl,wang2025stepsearch}. While early efforts successfully adapted classical algorithms like DQN for text-based games \cite{narasimhan2015language,mnih2015human}, subsequent research rapidly shifted toward policy gradient and value-based methods \cite{ammanabrolu2019playing,zha2021douzero}. Standard on-policy algorithms, such as PPO and AWR \cite{peng2019advantage}, proved effective in diverse interactive scenarios ranging from embodied tasks to complex card games. Despite these successes, efficiency and stability in training remain critical challenges. CoSo \cite{feng2025towards} addresses this with entropy-based regularization, while LOOP \cite{chen2025reinforcement} proposes a hybrid approach combining RLOO with PPO-style updates. Building on trajectory-level concepts, RAGEN \cite{wang2025ragen} utilizes StarPO to optimize entire interaction trajectories. However, RAGEN’s reliance on trajectory-level rewards prevents it from evaluating individual steps. GEPO \cite{yuan2025graph} combines graph centrality and task similarity to estimate task-conditioned state criticality for credit assignment. While GiGPO uses anchor state grouping to compare steps at identical states, it still estimates rewards based on single-trajectory outcomes, which suffers from sampling bias. In contrast, our approach aggregates identical states across all sampled trajectories into a state group graph, achieving fine-grained reward estimation with significantly lower variance.

\section{Preliminary}
\subsection{Problem Setup}
We consider a general scenario where an LLM agent solves a given task $x \in X$ through multi-turn, long-horizon interactions with an environment. At each time step $t=1,2,3 \dots, T$, the agent receives an observation $o_t$ (where $o_1=x$) from the environment and produces a new action $a_t$. The environment receives $a_t$ and then returns the subsequent observation $o_{t+1}$. This cycle yields a complete trajectory $\tau = (o_1, a_1, \dots, o_T ,a_T)$. The observation here incorporates a fixed number of historical steps. The agent's policy can be formulated as $\pi_\theta(a_i|o_i)$, where $\theta$ represents the model parameters. Only when the task completed or the interaction threshold reached, the environment returns a scalar reward $R$. Consequently, the agent receives no immediate rewards during intermediate steps, which makes scoring the agent's intermediate actions challenging.

\subsection{Group Relative Policy Optimization}

Recently, Group Relative Policy Optimization (GRPO) has served as a popular reinforcement learning algorithm that eliminates the computational burden of maintaining a separate value function critic. Instead, GRPO estimates the baseline using the mean reward of a group of sampled outputs. For each query $q$, GRPO generates a group of $G$ outputs $\{y_1, y_2, \dots, y_G\}$ from the old policy $\pi_{\theta_{\text{old}}}$. The advantage $A_i$ for the $i$-th output is computed by normalizing its reward $R_i$ against the group statistics: $A_i = \frac{R_i - \mu}{\sigma}$, where $\mu$ and $\sigma$ represent the mean and standard deviation of the rewards within the group, respectively. The objective function maximizes the clipped surrogate objective while incorporating a KL-divergence penalty to ensure training stability:

\begin{equation}
\resizebox{0.9\textwidth}{!}{$
    \mathcal{J}_{\text{GRPO}}(\theta) = \mathbb{E}_{\substack{q \sim \mathcal{D} \\ \{y_i\}_{i=1}^G \sim \pi_{\theta_{\text{old}}}}} \left[ \frac{1}{G} \sum_{i=1}^G \left( \min \left( \rho_i A_i, \text{clip}(\rho_i, 1-\epsilon, 1+\epsilon) A_i \right) - \beta \mathbb{D}_{\text{KL}}(\pi_\theta(\cdot|q) || \pi_{\text{ref}}(\cdot|q)) \right) \right]$
    },
\end{equation}

where $\rho_i = \frac{\pi_\theta(y_i|q)}{\pi_{\theta_{\text{old}}}(y_i|q)}$ denotes the importance sampling ratio, $\epsilon$ is the clipping parameter, and $\beta$ controls the strength of the regularization towards the reference policy $\pi_{\text{ref}}$.
\section{Training LLM Agents with \MethodName{}}
\label{sec:method}
\begin{figure}[!t]
\centering
\includegraphics[width=1.0\columnwidth]{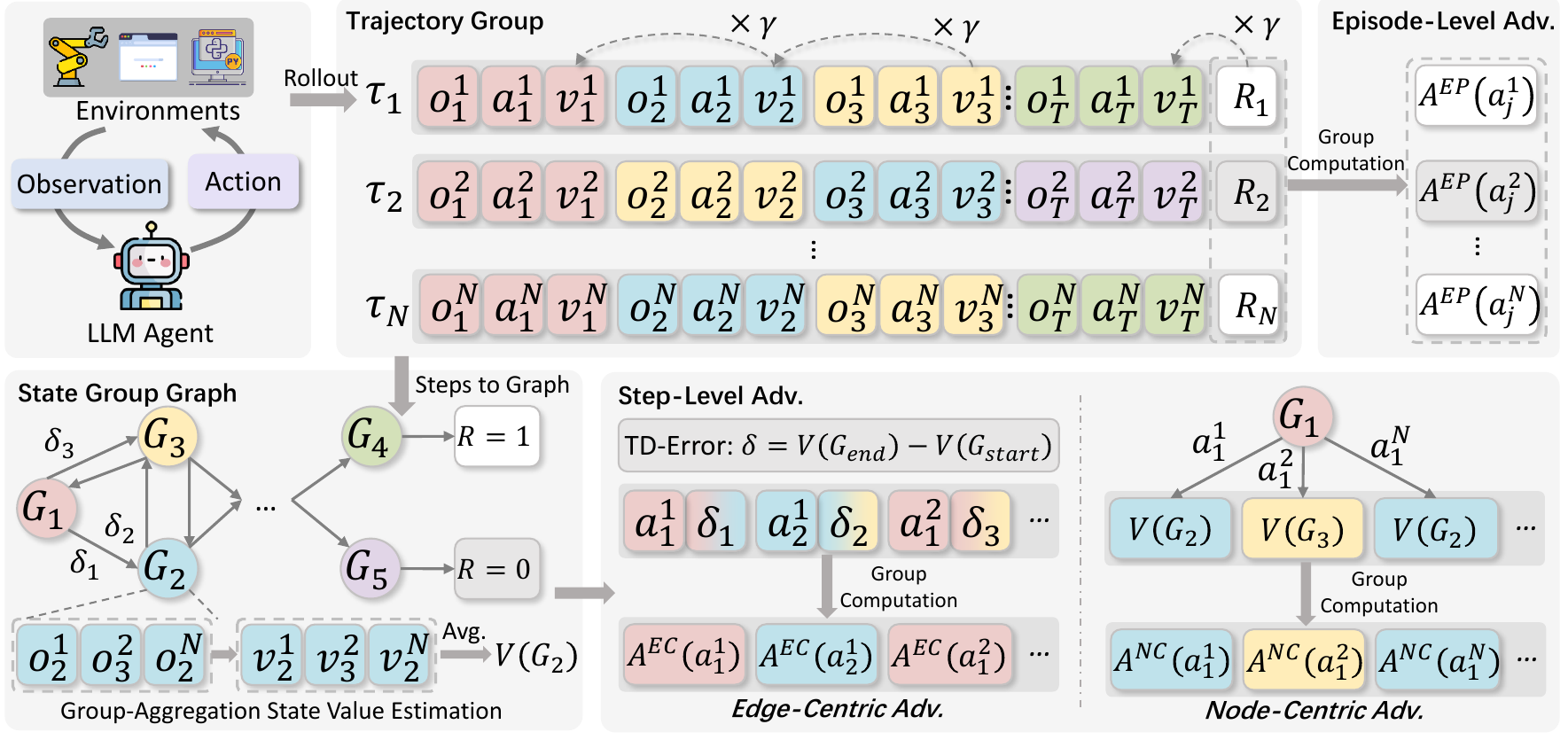}
\caption{Overview of \MethodName. For the sampled trajectories, we first construct a state group graph by aggregating steps with similar observations (same color) into the same state groups. We then define the state-group value as the average value of its constituent observations. Based on this graph, we provide step-level advantages from the edge and node perspectives, respectively.}
\label{fig:method}
\vspace{-0.5cm}
\end{figure}

In this section, we propose \MethodName{} to provide fine-grained credit assignment for multi-turn agentic RL from the perspective of a state group graph, as illustrated in Figure \ref{fig:method}. We first introduce the state graph construction strategy. We then present the state-value estimation method designed to mitigate sampling variance and finally detail the advantage calculation process.

\subsection{State Group Graph Construction}

In multi-turn agentic RL, the environment returns an observation in each step. The observation allows us to infer the corresponding environmental state and group the states across different sampled trajectories. For instance, in e-commerce scenario, identical web page content at different intermediate steps can be considered to belong to the same state group. Transitions between these state group nodes occur via the agent's actions, thereby forming the edges.

Formally, for a given task $x$, we first sample $N$ complete trajectories $\{\tau_i\}_{i=1}^{N}$ using the policy $\pi_{\theta_{old}}$, where $\tau_i=\{o_1^i,a_1^i,o_2^i,a_2^i,...,o_T^i,a_T^i\}$ and all trajectories satisfy $o_1^i=x$. Let $\mathcal{O} = \{o_j^i\}_{i=1, j=1}^{N, T}$ be the set of all intermediate observations. We then partition $\mathcal{O}$ into $G$ disjoint state groups $\{\mathcal{G}_k\}_{k=1}^G$ by clustering identical observations. Specifically, each group is defined as $\mathcal{G}_k = \{o_j^i \in \mathcal{O} \mid o_j^i = \bar{o}_k\}$, where $\bar{o}_k$ represents the unique observation characterizing the $k$-th state group. By treating these state groups as nodes and agent actions as edges, we construct a state transition graph $\{\mathcal{V},\mathcal{E}\}$, where $\mathcal{V}=\{\mathcal{G}_k\}_{k=1}^G$ and $\mathcal{E} = \{ (\mathcal{G}_s, a, \mathcal{G}_t) \mid \exists i, j \text{ s.t. } o_j^i \in \mathcal{G}_s, a_j^i = a, o_{j+1}^i \in \mathcal{G}_t \}$.

\subsection{Group-Aggregation State-Value Estimation}

After constructing the state graph, different from prior studies that directly use the reward of a single trajectory to calculate the value of intermediate states, we average step reward of all intermediate steps within the state node as the state value. Our motivation is twofold:
\vspace{-5pt}
\begin{itemize}[leftmargin=15pt, itemsep=0pt]
    \item Variance reduction: Due to sampling randomness, identical states in different trajectories may lead to different outcomes. Reliance on raw reward of a single sampling for value calculation introduces significant variance.
    \item Improved temporal credit assignment: In multi-turn RL, a good action may fail due to later errors, while a mediocre one succeeds by luck. Group aggregation averages out this future-step noise, ensuring more reliable and stable state evaluation.
\end{itemize}
\vspace{-5pt}
By averaging across different attempts passing through the same state node, we estimate the intrinsic expected value of the state, effectively mitigating value estimation variance, as formally proven in Appendix \ref{sec:value_var}. Specifically, we first independently calculate the state value for steps in each trajectory using discounted returns: 
\begin{equation}
    v_j^i=\gamma^{T-j+1} R_i,
    \label{eq:vij}
\end{equation}
where $\gamma \leq 1$ is the discount factor and $R_i$ is the outcome reward of $\tau_i$. Subsequently, for each state group $\mathcal{G}_k$, we define its value as the average of the values of all constituent steps:
\begin{equation}
    V(\mathcal{G}_k)=\frac{1}{|\mathcal{G}_k|}\sum\limits_{o_j^i\in \mathcal{G}_k} v_j^i. \label{eq:value}
\end{equation}

\subsection{Multi-Granularity Advantage Estimation}

In agentic RL, GRPO typically calculates advantages across trajectories based solely on the outcome reward. Steps within the corresponding trajectory share this advantage. This coarse approach can mislead the optimization, where good actions receive negative advantages due to subsequent failures, or mediocre actions receive positive advantages due to subsequent corrections. To address this, we leverage the state group graph to synthesize three granularities of advantage to provide dense and multi-granularity feedback.

\textbf{Node-Centric Advantage Estimation.} When the agent observes a specific environmental state $\bar o_k$, it may select from various actions leading to different subsequent states. These actions, originating from the same node, constitute a node-centric group. Formally, for any state group $\mathcal{G}_k$, we collect the values of all next-states resulting from transitions out of $\mathcal{G}_k$ to form a local reference set $\Delta_k = \{ V(\mathcal{G}) \mid o_m^n \in \mathcal{G}_k,\ o_{m+1}^n \in \mathcal{G} \}$. For any observation $o_j^i$ in $\mathcal{G}_k$, the node-centric advantage of corresponding action $a_j^i$is calculated as:
\begin{equation}
    A^{NC}(a_j^i) = \frac{V(\mathcal{G}_{k'}) - \text{mean}(\Delta_k)}{\text{std}(\Delta_k)}, \label{eq:anc}
\end{equation}

where next observation $o_{j+1}^i \in \mathcal{G}_k'$.

\textbf{Edge-Centric Advantage Estimation.} Within the state group graph, executing action $a_j^i$ transitions the environments from the current state group $\mathcal{G}_k$ ($o_j^i \in \mathcal{G}_k$) to the subsequent state group $\mathcal{G}_{k'}$ ($o_{j+1}^i \in \mathcal{G}_{k'}$). We quantify the quality of this transition by measuring its intrinsic value gain. In typical long-horizon agent tasks with sparse terminal rewards and an undiscounted horizon (i.e., $\gamma=1$), this value gain structurally corresponds to the 1-step TD error \cite{sutton2018reinforcement}, defined as:
\begin{equation}
    \delta_j^i=V(\mathcal{G}_{k'})-V(\mathcal{G}_k). \label{eq:delta}
\end{equation}

Standard node-centric methods normalize advantages locally among actions sharing the same state. However, local normalization obscures the absolute significance of a transition. A marginally better action in a trivial state might improperly receive the same advantage score as a critical breakthrough action. To capture the global significance of each transition, we propose treating the TD errors of all intermediate steps across all sampled trajectories as a global reference set. The edge-centric advantage is then computed via global standardization:
\begin{equation}
    A^{EC}(a_j^i)=\frac{\delta_j^i-\text{mean}(\{\delta_j^i\}_{i=1, j=1}^{N, T})}{\text{std}(\{\delta_j^i\}_{i=1, j=1}^{N, T})}. \label{eq:aec}
\end{equation}
The theoretical insight of this edge-centric formulation lies in its robust approach to the temporal credit assignment problem in sparse-reward RL. By globally standardizing TD errors across the entire state-transition graph, $A^{EC}$ explicitly identifies and prioritizes \textit{critical transitions}—actions that yield a substantial leap toward task completion. Consequently, an impactful action receives proportional positive reinforcement even if its specific trajectory ultimately fails due to subsequent exploratory errors. Conversely, trivial actions in successful trajectories are naturally penalized or down-weighted, effectively mitigating the noise inherent in long-horizon exploration. Meanwhile, this approach can reduce variance of advantage estimation compared to GRPO when the group size exceeds 1, as formally proven in Appendix \ref{sec:adv_var}.

\textbf{Episode-Level Advantage Estimation.} Although $A^{NC}$ and $A^{EC}$ provide dense and fine-grained feedback, it is crucial to ensure the learning process does not deviate from the ultimate task objective. Therefore, we retain an episode-level advantage similar to the original GRPO:
\begin{equation}
    A^{EP}(a_j^i)=\frac{R_i-\text{mean}(\{R_i\}_{i=1}^N)}{\text{std}(\{R_i\}_{i=1}^N)}, \label{eq:aep}
\end{equation}
where $R_i$ is the final reward for trajectory $\tau_i$.

Finally, we merge the episode-level and step-level advantage to compute the overall advantage for each step:

\begin{equation}
    A_j^i=A^{EP}(a_j^i)+w \cdot (A^{NC}(a_j^i)+A^{EC}(a_j^i)), \label{eq:aall}
\end{equation}
where $w$ is a static weight used to scale the contribution of the step-level advantage.

Following the GRPO framework, the overall objective function of \MethodName{} is formulated as:

\begin{equation}
\resizebox{0.93\textwidth}{!}{$
    \mathcal{J}(\theta) = \mathbb{E}_{\substack{x \sim p(X) \\ \{\tau_i\}_{i=1}^N \sim \pi_{\theta_{\text{old}}}}} \left[ \frac{1}{NT} \sum_{i=1}^{N} \sum_{j=1}^{T} \left( \min \left( \rho_j^i A_j^i, \operatorname{clip}\left( \rho_j^i, 1-\epsilon, 1+\epsilon \right) A_j^i \right) - \beta D_{\text{KL}} \left( \pi_\theta(\cdot|o_j^i) \| \pi_{\text{ref}}(\cdot|o_j^i) \right) \right) \right]$} \label{eq:mypo}
\end{equation}

where $\rho_j^i = \frac{\pi_\theta(a_j^i|o_j^i)}{\pi_{\theta_{old}}(a_j^i|o_j^i)}$ denotes the importance sampling ratio. $\epsilon$ is the clipping parameter to constrain policy updates, and $\beta$ is the coefficient for the KL-divergence penalty term $D_{KL}$.

\section{Experiment}
\subsection{Experimental Setting}
\label{sec:exp_setup}

\textbf{Benchmarks.} We conduct experiments on three challenging agentic benchmarks across different scenarios: WebShop \cite{yao2022webshop} for e-commerce, ALFWorld \cite{shridhar2020alfworld} for embodied physical environments, and AppWorld \cite{trivedi2024appworld} for complex software ecosystems. These benchmarks comprehensively assess the agents' long-horizon decision-making abilities in complex environments.

\textbf{Baselines.} We comprehensively compare \MethodName{} against three categories of competitive baselines: (1) Frontier off-the-shelf models, representing the upper bound of direct-inference capabilities, including Gemini-2.5-Pro \cite{comanici2025gemini}, DeepSeek-V3.2 \cite{liu2025deepseek}, and Qwen3.5-397B \cite{yang2025qwen3}; (2) Prompting agents, that leverage inference-time compute, including ReAct \cite{yao2022react} and Reflexion \cite{shinn2023reflexion}; (3) RL training methods, including PPO \cite{schulman2017proximal} as the standard actor-critic baseline, RLOO \cite{ahmadian2024back} which bypasses the critic model via a leave-one-out advantage baseline, GRPO \cite{shao2024deepseekmath} which normalizes rewards within sampled groups to reduce memory overhead, and GiGPO \cite{fenggroup} which introduces sub-grouping structures for fine-grained advantage estimation.

\textbf{Implementation details.} We primarily use Qwen2.5-1.5B-Instruct and Qwen2.5-7B-Instruct \cite{qwen2.5} as base models, while adopting Qwen2.5-14B-Instruct for AppWorld due to its high difficulty. All RL methods share the same hyperparameters and use identical prompts for rollouts, where <think></think> encapsulates the reasoning process and <action></action> specifies the executed action. We apply an invalid action penalty with a coefficient of 0.1 and set the learning rate to 1e-6. We use vLLM \cite{kwon2023efficient} as the inference engine. For group-based methods, the number of rollouts per task $N$ is set to 8 in all main experiments. Further implementation details can be found in Appendix \ref{sec:exp_detail}.

\textbf{State grouping.} All main-paper experiments group observations by exact string equality: webpage content (WebShop), local environment descriptions (ALFWorld), and Python API returns or errors (AppWorld). Exact matching is an implementation choice rather than an algorithmic requirement. \MethodName{} only requires an observation-to-node mapping; canonicalization or lexical/semantic similarity can replace exact matching without changing value estimation or optimization. Appendix~\ref{sec:noisy_grouping} details a noisy ALFWorld experiment validating this flexibility.

\subsection{Experimental Results}

\begin{table}[t]
\centering
\caption{Results on WebShop and ALFWorld. We report the success rate (\%) for each sub-task and overall results for ALFWorld. We report the score and success rate (\%) for WebShop. Results are averaged over 3 random seeds. The best results are highlighted in \textbf{bold}.}
\label{tab:main}
\resizebox{\textwidth}{!}{
\begin{tabular}{llccccccc|cc}
\toprule
\multirow{2}{*}{Type} & \multirow{2}{*}{Method} & \multicolumn{7}{c|}{\textbf{ALFWorld}} & \multicolumn{2}{c}{\textbf{WebShop}} \\
 & & Pick & Look & Clean & Heat & Cool & Pick2 & All & Score & Succ.\\
\midrule
\multicolumn{10}{l}{\textit{Off-the-Shelf Model}} \\
Prompting& Qwen3.5-397B & 96.6 & 72.7 & 70.4 & 0.0 & 70.4 & 85.7 & 71.9 & 27.6 & 17.2\\
Prompting& DeepSeek-V3.2 & 79.3 & 54.6 & 48.2 & 46.2 & 29.6 & 57.1 & 53.1 & 31.6 & 18.0\\
Prompting& Gemini-2.5-Pro & 92.8 & 63.3 & 62.1 & 69.0 & 26.6 & 58.7 & 60.3& 42.5 & 35.9\\
\midrule
\multicolumn{10}{l}{\textit{Qwen2.5-1.5B-Instruct}} \\
Prompting& Qwen2.5 & 5.9 & 5.5 & 3.3 & 9.7 & 4.2 & 0.0 & 4.1 & 23.1 & 5.2\\
Prompting& ReAct & 17.4 & 20.5 & 15.7 & 6.2 & 7.7 & 2.0 & 12.8& 40.1& 11.3\\
Prompting& Reflexion & 35.3 & 22.2 & 21.7 & 13.6 & 19.4 & 3.7 & 21.8 & 55.8& 21.9\\
RL Training& PPO (with critic) & 64.8\textsubscript{\textpm3.5} & 40.5\textsubscript{\textpm6.9} & 57.1\textsubscript{\textpm4.9} & 60.6\textsubscript{\textpm6.6} & 46.4\textsubscript{\textpm4.0} & 47.4\textsubscript{\textpm1.9} & 54.4\textsubscript{\textpm3.1}& 73.8\textsubscript{\textpm3.0} & 51.5\textsubscript{\textpm2.9} \\
RL Training& RLOO & 88.3\textsubscript{\textpm3.0} & 52.8\textsubscript{\textpm8.6} & 71.0\textsubscript{\textpm5.9} & 62.8\textsubscript{\textpm8.7} & 66.4\textsubscript{\textpm5.5} & 56.9\textsubscript{\textpm4.7} & 69.7\textsubscript{\textpm2.5}& 73.9\textsubscript{\textpm5.6}& 52.1\textsubscript{\textpm6.7}\\
RL Training& GiGPO & 94.4\textsubscript{\textpm5.9} & 67.5\textsubscript{\textpm4.6} & 94.8\textsubscript{\textpm3.8} & 94.4\textsubscript{\textpm7.8} & 79.8\textsubscript{\textpm4.7} & 76.4\textsubscript{\textpm5.4} & 86.7\textsubscript{\textpm1.7} & 83.5\textsubscript{\textpm1.8}& 67.4\textsubscript{\textpm4.5}\\
RL Training& GRPO & 85.3\textsubscript{\textpm1.5} & 53.7\textsubscript{\textpm8.0} & 84.5\textsubscript{\textpm6.8} & 78.2\textsubscript{\textpm7.9} & 59.7\textsubscript{\textpm5.0} & 53.5\textsubscript{\textpm5.6} & 72.8\textsubscript{\textpm3.6}& 75.8\textsubscript{\textpm3.5} & 56.8\textsubscript{\textpm3.8}\\
\rowcolor{gray!15}RL Training& \textbf{\MethodName} & \textbf{97.2}\textsubscript{\textpm3.9} & \textbf{96.3}\textsubscript{\textpm5.2} & \textbf{95.1}\textsubscript{\textpm1.8} & \textbf{94.9}\textsubscript{\textpm3.6} & \textbf{97.1}\textsubscript{\textpm2.0} & \textbf{91.4}\textsubscript{\textpm4.1} & \textbf{95.0}\textsubscript{\textpm0.8} & \textbf{85.1}\textsubscript{\textpm1.4} & \textbf{71.2}\textsubscript{\textpm2.6}\\
\rowcolor{gray!15}RL Training& $\Delta$ \textit{vs GRPO} & \textit{+11.9} & \textit{+42.6} & \textit{+10.6} & \textit{+16.7} & \textit{+37.4} & \textit{+37.9} & \textit{+22.2} & \textit{+9.3} & \textit{+14.4}\\
\midrule
\multicolumn{10}{l}{\textit{Qwen2.5-7B-Instruct}} \\
Prompting& Qwen2.5 & 33.4 & 21.6 & 19.3 & 6.9 & 2.8 & 3.2 & 14.8 & 26.4 & 7.8\\
Prompting& ReAct & 48.5 & 35.4 & 34.3 & 13.2 & 18.2 & 17.6 & 31.2 & 46.2 & 19.5\\
Prompting& Reflexion & 62.0 & 41.6 & 44.9 & 30.9 & 36.3 & 23.8 & 42.7& 58.1& 28.8\\
RL Training& PPO (with critic) & 92.3\textsubscript{\textpm4.0} & 64.0\textsubscript{\textpm8.4} & 92.5\textsubscript{\textpm2.4} & 89.5\textsubscript{\textpm7.0} & 80.3\textsubscript{\textpm2.0} & 68.8\textsubscript{\textpm8.3} & 80.4\textsubscript{\textpm2.7} & 81.4\textsubscript{\textpm3.1}& 68.7\textsubscript{\textpm5.1}\\
RL Training& RLOO & 87.6\textsubscript{\textpm4.3} & 78.2\textsubscript{\textpm8.3} & 87.3\textsubscript{\textpm5.8} & 81.3\textsubscript{\textpm7.6} & 71.9\textsubscript{\textpm5.2} & 48.9\textsubscript{\textpm8.4} & 75.5\textsubscript{\textpm4.6} & 80.3\textsubscript{\textpm3.2} & 65.7\textsubscript{\textpm4.0}\\
RL Training& GiGPO & 97.7\textsubscript{\textpm1.6} & 82.7\textsubscript{\textpm7.9} & \textbf{98.8}\textsubscript{\textpm1.6} & 83.7\textsubscript{\textpm7.2} & 89.3\textsubscript{\textpm8.2} & 79.2\textsubscript{\textpm6.6} & 90.8\textsubscript{\textpm1.3} & 86.2\textsubscript{\textpm2.6} & 75.2\textsubscript{\textpm3.8}\\
RL Training& GRPO & 90.8\textsubscript{\textpm5.1} & 66.1\textsubscript{\textpm6.7} & 89.3\textsubscript{\textpm5.4} & 74.7\textsubscript{\textpm6.9} & 72.5\textsubscript{\textpm5.4} & 64.7\textsubscript{\textpm7.3} & 77.6\textsubscript{\textpm5.2} & 79.3\textsubscript{\textpm2.8} & 66.1\textsubscript{\textpm3.7}\\ 
\rowcolor{gray!15}RL Training& \textbf{\MethodName} & \textbf{98.7}\textsubscript{\textpm2.3} & \textbf{90.6}\textsubscript{\textpm11.5} & 97.6\textsubscript{\textpm2.1} & \textbf{100.0}\textsubscript{\textpm0.0} & \textbf{98.8}\textsubscript{\textpm2.1} & \textbf{93.7}\textsubscript{\textpm8.4} & \textbf{96.9}\textsubscript{\textpm1.3} & \textbf{89.8}\textsubscript{\textpm1.7} & \textbf{78.3}\textsubscript{\textpm0.6}\\
\rowcolor{gray!15}RL Training& $\Delta$ \textit{vs GRPO} & \textit{+7.9} & \textit{+24.5} & \textit{+8.3} & \textit{+25.3} & \textit{+26.3} & \textit{+29} & \textit{+19.3} & \textit{+10.5} & \textit{+12.2}\\
\bottomrule
\end{tabular}
}
\end{table}

\textbf{Performance on WebShop and ALFWorld}. As shown in Table \ref{tab:main}, even frontier LLMs fall short on these two benchmarks. For example, Qwen3.5-397B yields only a 17.2\% success rate on WebShop. Moreover, RL training methods significantly outperform prompt-based methods, highlighting their critical role in internalizing agent capabilities into the model. Among these RL methods, \MethodName{} consistently surpasses all baselines at both 1.5B and 7B scales. In particular, \MethodName{} achieves improvements with a substantial margin of up to 22\% and 14\% over the GRPO baseline in WebShop and ALFWorld, respectively. Furthermore, compared to GiGPO which relies on comparison within local groups, \MethodName{} still performs better, demonstrating the importance of edge-centric advantage which offers global signals.

\begin{wraptable}{r}{0.4\textwidth}
\vspace{-10pt}
\centering
\caption{Results on AppWorld.}
\label{tab:appworld}
\resizebox{0.4\textwidth}{!}{\begin{tabular}{llcc}
\toprule
Type & Method & Succ. & Score \\
\midrule
\multicolumn{4}{l}{\textit{Qwen2.5-14B-Instruct}} \\
Prompting& Qwen2.5 & 0.0 & 0.0\\
Prompting& ReAct & 10.5 & 8.0\\
RL Training& PPO (with critic)  & 19.1 & 14.7\\
RL Training& RLOO & 24.8 & 19.5\\
RL Training& GiGPO & 25.7 & 19.2\\
RL Training& GRPO &24.8 & 20.7\\
\rowcolor{gray!15}RL Training& \MethodName & \textbf{27.6} & \textbf{21.7}\\
\bottomrule
\end{tabular}}
\end{wraptable}

\textbf{Performance on AppWorld}. As shown in Table \ref{tab:appworld}, \MethodName{} outperforms all RL training methods, proving its effectiveness and strong generalizability even on a more challenging benchmark like AppWorld. In AppWorld, agents interact with environments via Python API calls. This requires the agent to infer the app's underlying data state from specific API returns and effectively execute subsequent actions based on this state. Despite such complex observations, G2PO can still construct state transition graphs and provide fine-grained credit assignment to distinguish valid from invalid actions. This highlights G2PO's robust generalizability across diverse observation modalities.

\subsection{Ablation Study}

\begin{wrapfigure}{R}{0.28\textwidth}
  \centering
  \includegraphics[width=\linewidth]{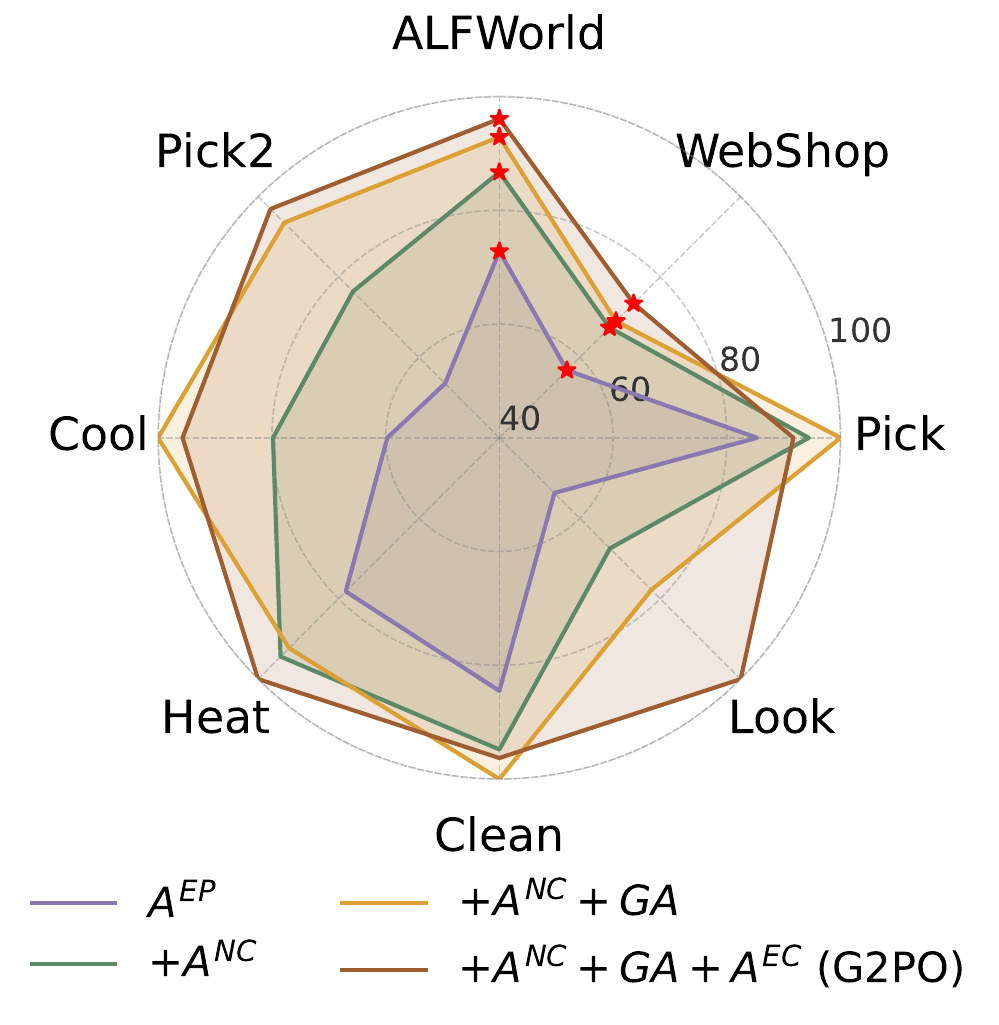}
  \vspace{-10pt}
  \caption{Ablation results. The y-axis shows success rate (\%).}
  \label{fig:ablation}
  \vspace{-10pt}
\end{wrapfigure}

In this part, we conduct ablation studies to validate the key components of \MethodName. We start with a baseline Episode-level Advantage ($A^{EP}$) and incrementally introduce Node-Centric Advantage ($A^{NC}$), Group-Aggregation (GA) state-value estimation, and Edge-Centric Advantage ($A^{EC}$) to evaluate their individual contributions. We train on Qwen2.5-1.5B-Instruct for ablation results.

As illustrated in Figure \ref{fig:ablation}, integrating the group-aggregation state-value estimation yields significant performance gains on both WebShop and ALFWorld. This improvement highlights its vital role in reducing the variance of state-value estimation by averaging the outcomes of identical states encountered across multiple trajectories. Furthermore, incorporating edge-centric advantage estimation outperforms relying solely on the node-centric approach. This confirms that the edge-centric method successfully captures the global, absolute contribution of actions, which local comparisons within a state group fail to reflect. Overall, the ablation results demonstrate that every component of \MethodName{} is indispensable for providing accurate and fine-grained advantage. Appendix~\ref{sec:edge_diagnostics} further examines whether edge-centric advantages assign greater credit to critical actions and correlate with action importance.

\subsection{Analysis}

\begin{figure}[t]
  \begin{subfigure}[b]{0.28\textwidth}
    \centering
    \includegraphics[width=\textwidth]{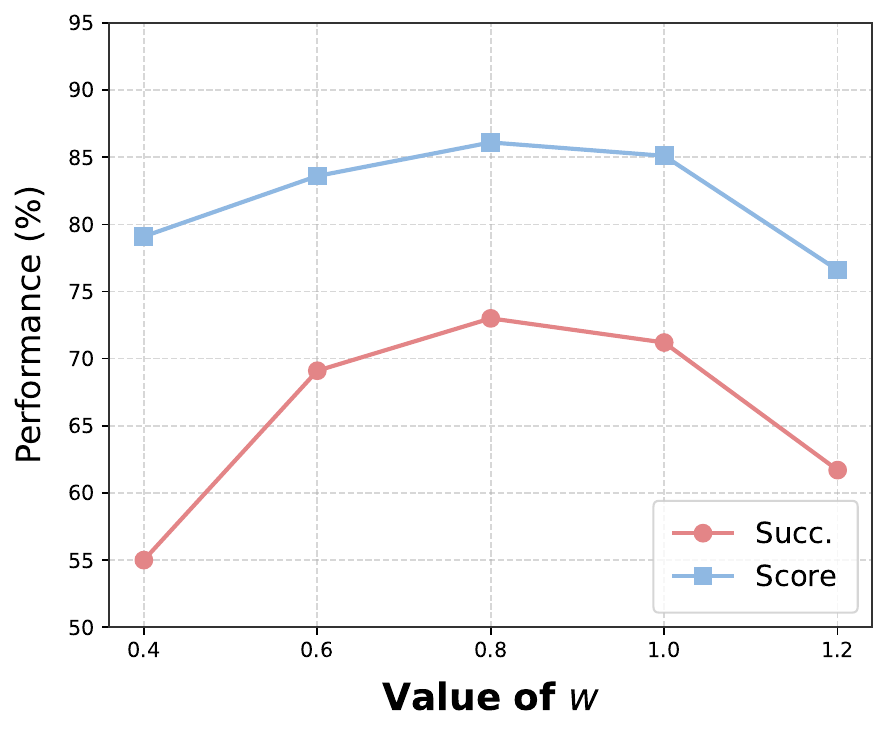}
    \caption{}
    \label{fig:hyper}
  \end{subfigure}
  \hfill 
  \begin{subfigure}[b]{0.3\textwidth}
    \centering
    \includegraphics[width=\textwidth]{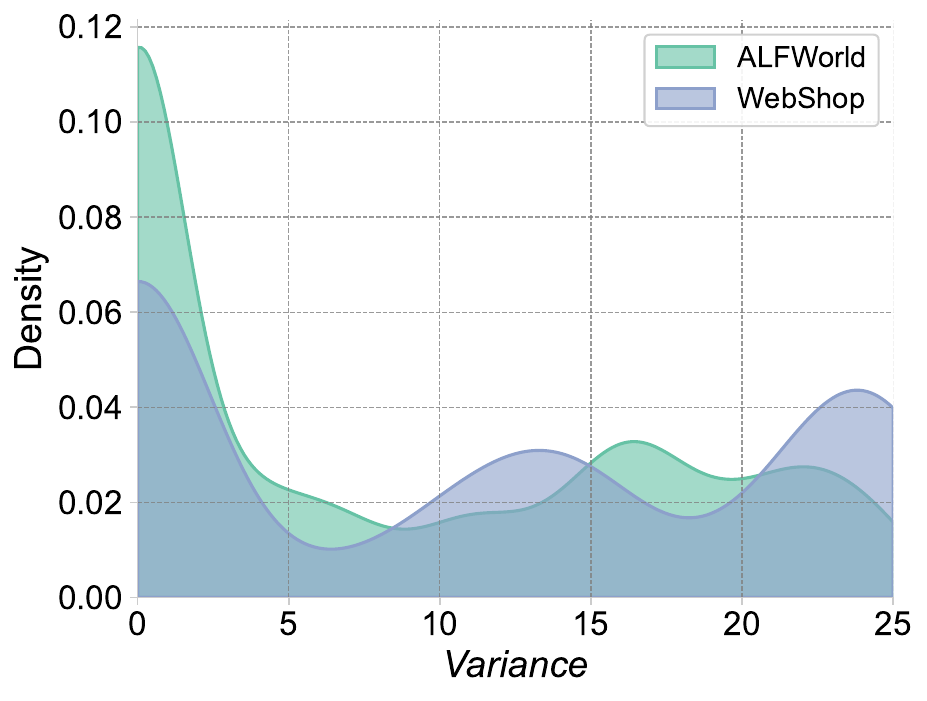}
    \caption{}
    \label{fig:var}
  \end{subfigure}
  \hfill
  \begin{subfigure}[b]{0.39\textwidth}
    \centering
    \includegraphics[width=\textwidth]{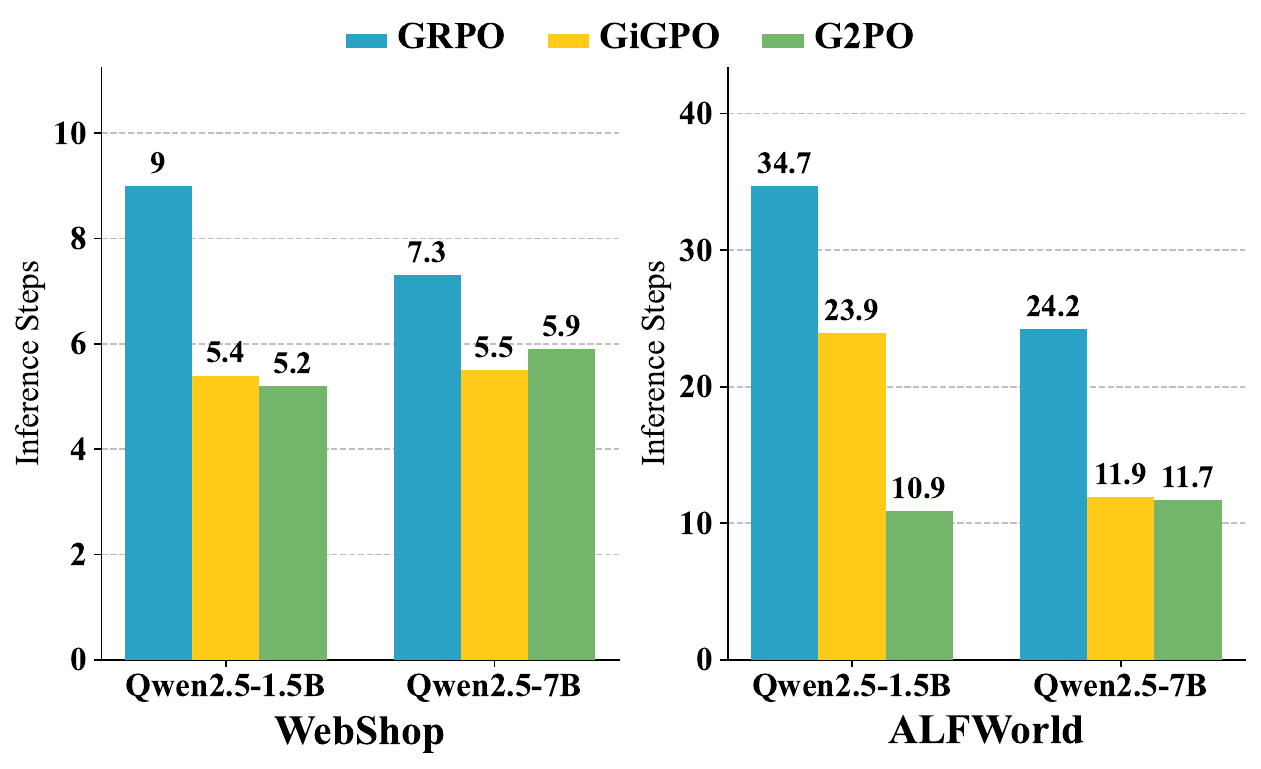}
    \caption{}
    \label{fig:infer}
  \end{subfigure}

  \caption{Results for analysis. Figure \ref{fig:hyper}: results under different values of parameter $w$; Figure \ref{fig:var}: variance distribution of state-value estimation with step-level RL methods, highlighting the necessity of group-aggregation mechanism to reduce variance; Figure \ref{fig:infer}: the number of inference steps required to finish tasks. \MethodName{} demonstrates substantial advantages over existing methods in inference efficiency.}
  \label{fig:analysis}
\end{figure}

\textbf{Hyperparameter Analysis.} Here we analyze the parameter $w$ in Eq. \ref{eq:aall}, which balances step-level and episode-level advantages. We conduct experiments using Qwen2.5-1.5B on WebShop to evaluate its impact. As shown in Figure \ref{fig:hyper}, the model's performance first increases and then decreases as $w$ grows, achieving the highest success rate and score at $w=0.8$. This highlights the parameter's critical role: when $w$ is too low, the algorithm degenerates to GRPO and fails to provide fine-grained advantages for individual steps; when $w$ is too high, the algorithm becomes overly short-sighted, causing the optimization to easily deviate from the ultimate task objective.

\textbf{State-Value Variance Analysis.} To illustrate the necessity of the group-aggregation mechanism in \MethodName{}, we present the variance statistics of value estimation with current step-level RL methods. We distinguish states based on environmental observations and collect the final task reward derived from a state as its state value. As illustrated in Figure \ref{fig:var}, across both benchmarks and 1.5B/7B model scales, a significant portion of states exhibits a non-zero value variance. This implies that starting from the same state, the agent often reaches different outcome states with varying rewards, resulting in inconsistent state values for identical states across different trajectories. To resolve this high-variance issue, our group-aggregation mechanism averages the diverse outcomes stemming from identical states across trajectories, assigning a robust and consistent value to each state node.

\textbf{Task Efficiency Analysis.} As shown in Figure \ref{fig:infer}, we collected the number of environment interaction turns required by the trained models to complete tasks. It is evident that, compared to GRPO, models trained with \MethodName{} complete tasks in fewer interaction turns on both WebShop and ALFWorld. In most cases, they also require fewer turns than GiGPO, indicating that \MethodName{} enhances the model's ability to make efficient decisions. Fewer interaction turns inherently translate to lower model inference costs and environment API call expenses. Consequently, \MethodName{} demonstrates substantial advantages over existing methods in both task accuracy and cost efficiency.

\subsection{Computational Budget}

\begin{wrapfigure}{l}{0.38\textwidth}
  \centering
  \includegraphics[width=\linewidth]{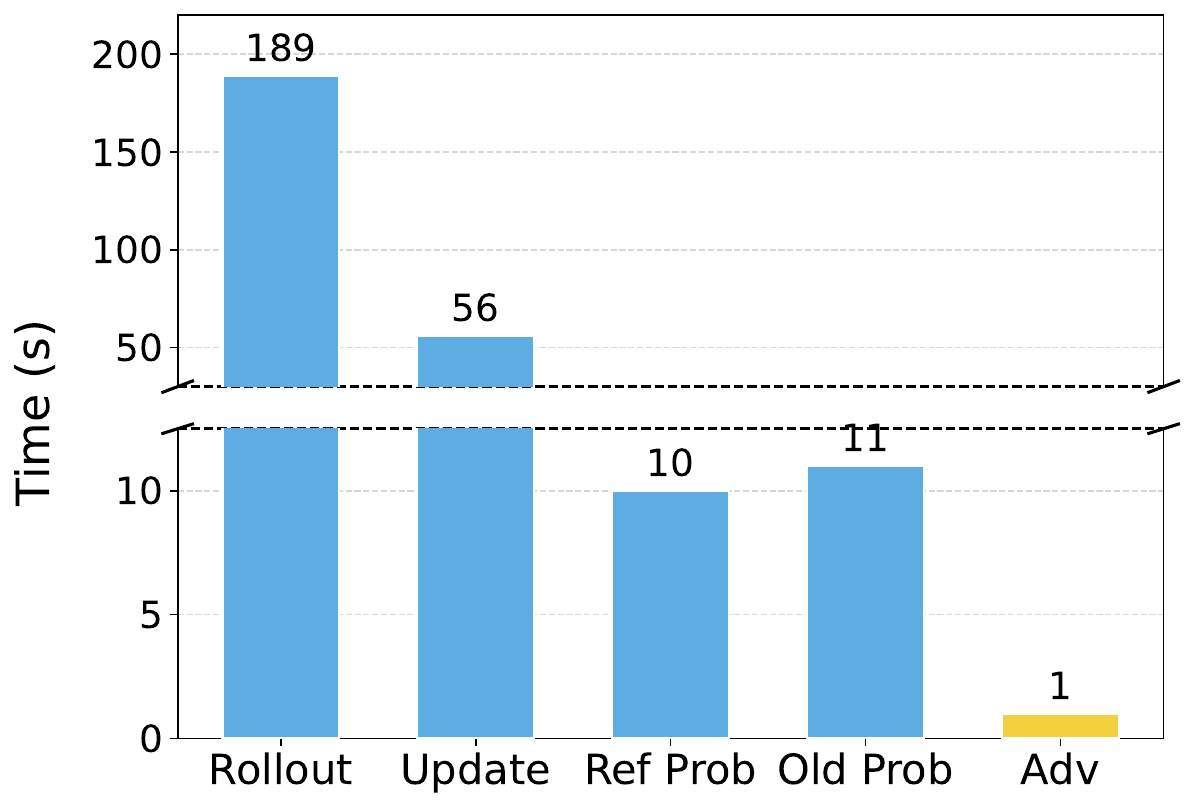}
  \vspace{-10pt}
  \caption{Breakdown of time consumption per training step. \MethodName{} introduces only 1s of additional time overhead when computing advantages.}
  \label{fig:compute}
  \vspace{-10pt}
\end{wrapfigure}

In this section, we demonstrate the efficiency of \MethodName{} by analyzing the time consumption per training step. We use Qwen2.5-1.5B-Instruct as the base model and train it on the ALFWorld benchmark.

As illustrated in Figure \ref{fig:compute}, a typical training step involves a multi-turn sampling process, calculating the output probabilities of the sampled responses under both the reference and old policies, estimating advantages, and updating the policy. \MethodName{} operates specifically within the advantage estimation stage. We can see from the result that \MethodName{} introduces only 1 second of additional overhead during advantage calculation, successfully preserving the high efficiency of GRPO. It is mainly due to its GPU-free design. Specifically, to estimate advantages, \MethodName{} constructs a state graph based on environmental observations, performs value estimation via group aggregation, and computes advantages at three different granularities. As this entire pipeline relies exclusively on the CPU, it occupies a mere 0.4\% of the total RL training time, introducing virtually no extra computational cost.

\section{Conclusion}
In this paper, we propose \MethodName{}, a novel group-based reinforcement learning framework for long-horizon agent tasks. By restructuring isolated interaction trajectories into a unified state-transition graph, \MethodName{} unlocks two key innovations: (i) \textit{group-aggregation state-value estimation}, which mitigates the high variance of single-trajectory outcomes, and (ii) \textit{edge-centric advantage estimation}, which evaluates the intrinsic value gain of each transition to globally prioritize critical breakthroughs. Experiments demonstrate that \MethodName{} achieves substantial performance gains over state-of-the-art RL baselines with almost no additional computational overhead.

\begin{ack}
\textbf{Funding.}
This work was supported by computing resources provided by Microsoft.

\textbf{Competing interests.}
The authors declare no competing financial interests.
\end{ack}

\bibliographystyle{plain}
\bibliography{reference.bib}

\medskip

\appendix
\section{Limitations}
\label{sec:limitations}
One limitation of our paper is that we only implement our method on the verl-agent \cite{fenggroup} framework, considering that it provides lightweight step-level RL and flexible credit assignment control. Experiments on more agentic RL frameworks will further demonstrate the usefulness of \MethodName{}. Another direction worthy of exploration is extending \MethodName{} to more rollout settings, such as tree sampling and asynchronous sampling.

\section{Variance Reduction Analysis}
\label{sec:proof}
In this section, we provide theoretical analysis and proof for the reduction of variance in value estimation through group-aggregation state-value estimation, as well as the reduction of variance in advantage estimation through edge-centric advantage estimation compared to trajectory-level credit assignment.
\subsection{Value Estimation Variance Reduction}
\label{sec:value_var}
In long-horizon agent tasks with sparse terminal rewards and an undiscounted horizon ($\gamma=1$), intermediate steps yield zero immediate rewards (i.e., $r_t=0$ for $t < T$). Consequently, the cumulative empirical return from any intermediate step $j$ precisely collapses to the final terminal reward $R_i$:
    $$v_j^i = \sum_{t=j}^T \gamma^{t-j} r_t = 0 + 0 + \dots + R_i = R_i.$$
Let $Var(R_i)=\sigma^2$. We have:
\begin{itemize}[leftmargin=15pt, itemsep=0pt]
    \item \textbf{Variance of Single-Trajectory Estimation:} In previous studies utilizing single-trajectory returns, the variance of the estimate is:
    $$\text{Var}(v_j^i) = \text{Var}(R_i) = \sigma^2.$$

    \item \textbf{Variance of \MethodName{} (State Group Aggregation):} For convenience, we assume that the sampling across different trajectories $\{\tau_i\}$ is independent and each intermediate step within $\mathcal{G}_k$ is sampled from a different trajectory. Then the variance of our aggregated estimate is:
    $$\text{Var}(V(\mathcal{G}_k)) = \text{Var}\left( \frac{1}{|\mathcal{G}_k|} \sum\limits_{o_j^i \in \mathcal{G}_k} v_j^i \right)= \frac{1}{|\mathcal{G}_k|^2} \sum\limits_{o_j^i \in \mathcal{G}_k} \text{Var}(R_i)= \frac{1}{|\mathcal{G}_k|^2} \cdot |\mathcal{G}_k| \cdot \sigma^2= \frac{\sigma^2}{|\mathcal{G}_k|}.$$
\end{itemize}

Consequently, when the state group size $|\mathcal{G}_k| > 1$, the variance of the value estimation decreases linearly relative to the single-trajectory estimate.
\subsection{Advantage Estimation Variance Reduction}
\label{sec:adv_var}
Here, we follow the assumptions used in the proof of the previous subsection.
\begin{itemize}[leftmargin=15pt, itemsep=0pt]
    \item \textbf{Variance of Traditional Trajectory-Level Advantage:} The trajectory-level advantage can be essentially viewed as $A^{traj}(a_j^i) = R_i - b$ (where $b$ is a scalar baseline). The variance of this estimation is entirely determined by the single-sample trajectory outcome:
    $$\text{Var}(A^{traj}(a_j^i)) = \text{Var}(R_i) = \sigma^2.$$
    This implies that trajectory-level advantage estimation is heavily influenced by the outcomes of subsequent steps, thereby failing to accurately assess the intrinsic expected advantage of the action itself.
    \item \textbf{Variance Reduction in Edge-Centric Advantage:} In contrast, \MethodName{} evaluates the action via the edge-centric advantage, which relies on the 1-step Temporal Difference (TD) error between the target and source state groups: $$\delta_j^i = V(\mathcal{G}_{k'}) - V(\mathcal{G}_k).$$ According to the fundamental properties of variance, this expands to:
    $$\text{Var}(\delta_j^i) = \text{Var}(V(\mathcal{G}_{k'}) - V(\mathcal{G}_k)) = \text{Var}(V(\mathcal{G}_{k'})) + \text{Var}(V(\mathcal{G}_k)) - 2\text{Cov}(V(\mathcal{G}_{k'}), V(\mathcal{G}_k)).$$
    Since trajectories passing through the target node $\mathcal{G}_{k'}$ inherently share historical context or overlap with those passing through the source node $\mathcal{G}_k$, their value estimates are positively correlated, i.e., $\text{Cov}(V(\mathcal{G}_{k'}), V(\mathcal{G}_k)) \ge 0$. Combining this with the variance of the group-aggregated state value derived previously, we obtain an upper bound for the variance of the edge-centric advantage:
    $$\text{Var}(\delta_j^i) \le \text{Var}(V(\mathcal{G}_{k'})) + \text{Var}(V(\mathcal{G}_k)) = \frac{\sigma^2}{|\mathcal{G}_{k'}|} + \frac{\sigma^2}{|\mathcal{G}_k|}.$$ Thus, when $|\mathcal{G}_k| > 1$ and $|\mathcal{G}_{k'}| > 1$, we have $$\text{Var}(\delta_j^i) \le \sigma^2.$$
    Consequently, the Edge-Centric Advantage formulation guarantees that, under these assumptions, the estimation variance is upper-bounded by the variance of trajectory-level advantage estimation.
\end{itemize}

\section{Pseudo Code}

Algorithm~\ref{alg:method} summarizes the full \MethodName{} training procedure. Compared to vanilla GRPO, we highlight the additional parts introduced by \MethodName{} in italics. In particular, constructing the state group graph and estimating state values through the group-aggregation mechanism mitigate sampling variance and trajectory-dependent bias. Furthermore, computing edge-centric and node-centric advantages provides multi-granularity feedback, which are adaptively integrated with episode-level advantages using confidence-aware weights. As such, \MethodName{} provides dense and fine-grained credit assignment while ensuring the learning process aligns with the ultimate task objective.

\begin{algorithm}[h]
\caption{Training LLM Agents with \MethodName{}}
\label{alg:method}
\begin{algorithmic}[1]
\STATE {\bfseries Require:} Initial policy $\pi_{\theta_{\text{old}}}$, reference policy $\pi_{\text{ref}}$, task distribution $p(X)$, discount factor $\gamma$, balancing weight $w$, clipping parameter $\epsilon$, KL penalty $\beta$, sample size $N$, maximum steps $T$
\FOR{each training iteration}
    \STATE Update the old policy model: $\theta_{\text{old}} \leftarrow \theta$
    \STATE \small{\color{gray}{// Multi-step rollout phase}}
    \STATE Sample task $x \sim p(X)$ and initialize $N$ identical environments
    \FOR{$j = 1$ to $T$}
        \STATE Sample actions $\bigl\{a_j^i \sim \pi_{\theta_{\text{old}}}(\cdot \mid o_j^i)\bigr\}_{i=1}^N$ (with $o_1^i = x$)
        \STATE Execute actions, observe next observations $\{o_{j+1}^i\}_{i=1}^N$
    \ENDFOR
    \STATE Collect final outcome rewards $\{R_i\}_{i=1}^N$ for each trajectory $\tau_i$

    \STATE \small{\color{gray}{// State group graph construction phase}}
    \STATE \emph{Partition intermediate observations $\mathcal{O}$ into $G$ state groups $\{\mathcal{G}_k\}_{k=1}^G$}
    \STATE \emph{Construct state transition graph $\{\mathcal{V},\mathcal{E}\}$ with state groups as nodes and actions as edges}
    
    \STATE \small{\color{gray}{// Group-aggregation state-value estimation phase}}
    \STATE \emph{Calculate discounted step values $v_j^i=\gamma^{T-j+1} R_i$ for all intermediate steps (Eq. \ref{eq:vij})}
    \STATE \emph{Estimate state group values $V(\mathcal{G}_k)$ by averaging step values within each group (Eq. \ref{eq:value})}
    
    \STATE \small{\color{gray}{// Multi-granularity advantage estimation phase}}
    \STATE \emph{Compute group state-value differences $\delta_j^i$ (Eq. \ref{eq:delta})}
    \STATE \emph{Compute edge-centric advantages $A^{EC}(a_j^i)$ via group state-value differences $\delta_j^i$ (Eq. \ref{eq:aec})}
    \STATE \emph{Compute node-centric advantages $A^{NC}(a_j^i)$ via local reference sets $\Delta_k$ (Eq. \ref{eq:anc})}
    \STATE Compute episode-level advantages $A^{EP}(a_j^i)$ via outcome rewards $R_i$ (Eq. \ref{eq:aep})
    
    \STATE \small{\color{gray}{// Policy update phase}}
    \STATE \emph{Combine advantages: $A_j^i=A^{EP}(a_j^i)+ w \cdot \left(A^{EC}(a_j^i)+A^{NC}(a_j^i)\right)$ (Eq. \ref{eq:aall})}
    \STATE Update policy $\theta$ by maximizing objective $\mathcal{J}(\theta)$ (Eq. \ref{eq:mypo})

\ENDFOR
\end{algorithmic}
\end{algorithm}

\section{Group Size Analysis}
\label{sec:group_size}
\begin{figure}[t] 
\centering
\begin{minipage}[t]{0.48\linewidth} 
\centering 
    \includegraphics[width=1.0\columnwidth]{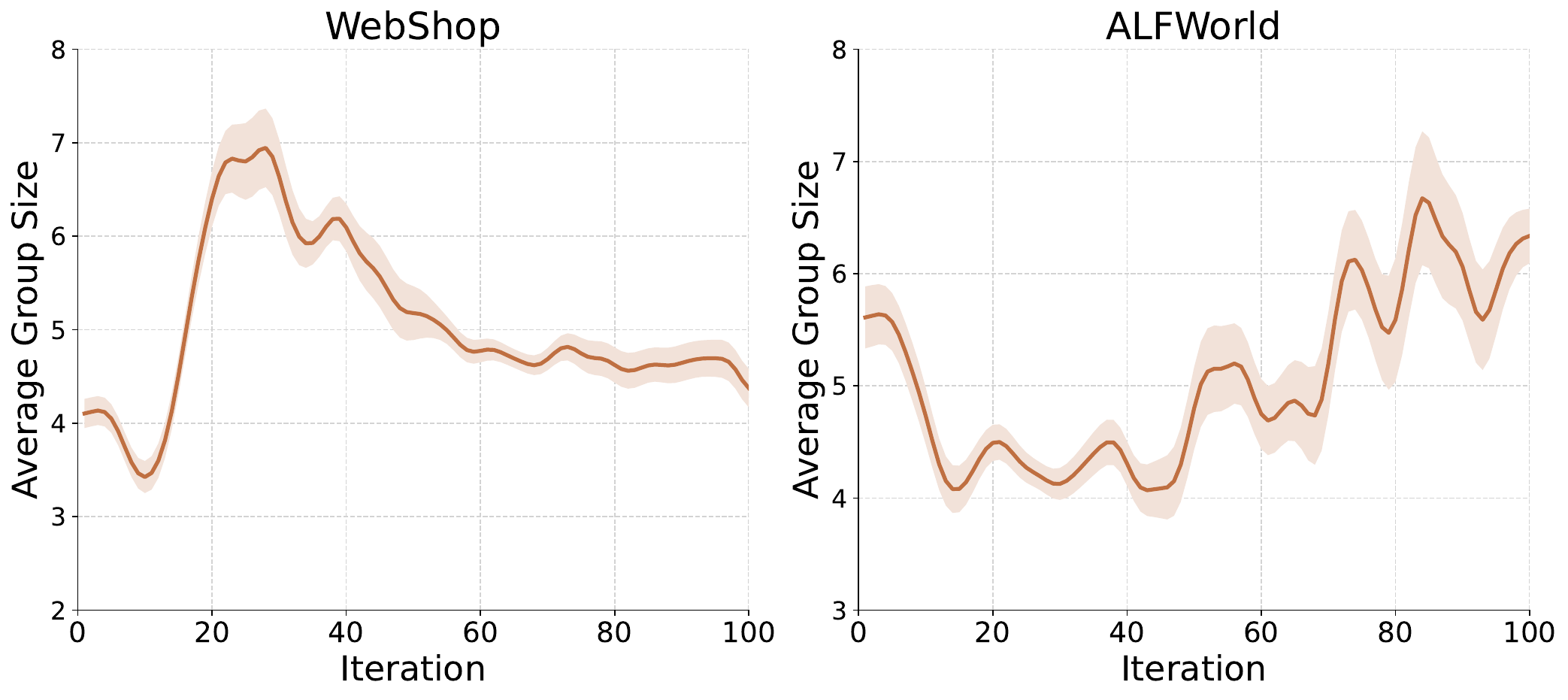}
    \vspace{-0.4cm}
    \caption{Dynamics of average group size during the training using Qwen2.5-1.5B-Instruct.}
    \label{fig:group_size}
\end{minipage} 
\hfill
\begin{minipage}[t]{0.50\linewidth} 
\centering 
    \includegraphics[width=1.0\columnwidth]{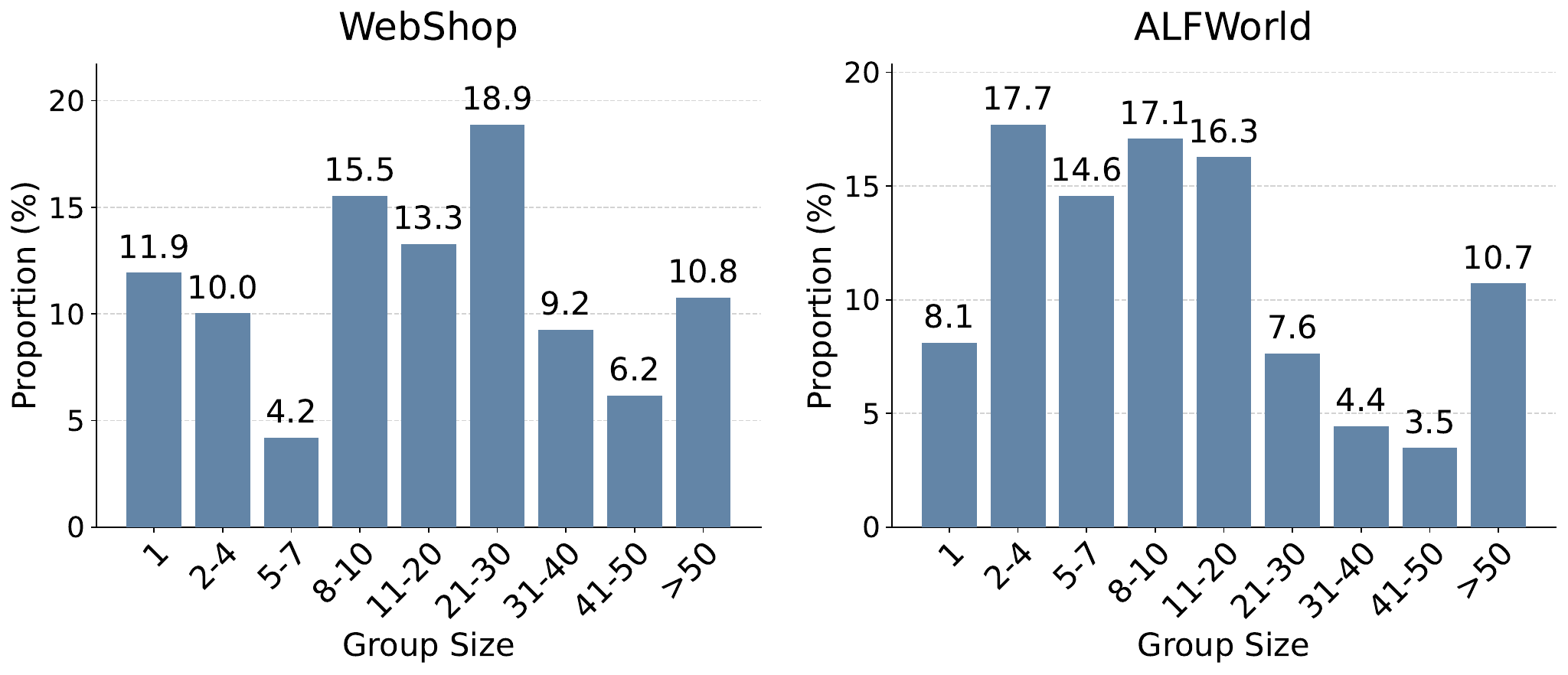}
    \vspace{-0.4cm}
    \caption{The distributions of group size using Qwen2.5-1.5B-Instruct}
    \label{fig:distribution}
\end{minipage} 
\end{figure}

Here we present the dynamics of the average state group size during training and the distribution of group sizes. As illustrated in Figure \ref{fig:group_size}, the average group size for both WebShop and ALFWorld is around 5. This indicates that many environmental observations are identical during training but may lead to different outcomes, highlighting the necessity of the group aggregation mechanism. In WebShop, the average group size initially increases, suggesting that the agent's decisions converge during early training, leading to identical observations. Subsequently, the group size decreases because fine-grained credit assignment helps the agent identify and eliminate redundant steps; this leads to shorter interaction trajectories and fewer total observations. Conversely, in ALFWorld, the average group size increases after the reasoning format learning phase. While training still drives decision convergence, ALFWorld tasks inherently require repetitive actions, forcing the agent to revisit identical states. This inherent repetition amplifies the upward trend in group size, outweighing the downward trend caused by redundant step removal. Furthermore, as shown in Figure \ref{fig:distribution}, it is evident that only 11.9\% and 8.1\% of steps in WebShop and ALFWorld, respectively, have a group size of 1. All remaining steps fall into groups of 2 or larger, meaning they all benefit from the reduced value estimation variance enabled by the group-aggregation mechanism.

For the more complex AppWorld benchmark, the mean state group size is 2.01 and the mean number of state groups per task graph is 155.38. Thus, overlap is lower than in WebShop and ALFWorld but remains non-trivial even in a long-horizon, API-based environment.

\section{Experiment Details}
\label{sec:exp_detail}

\subsection{Descriptions of Benchmarks}
\label{sec:benchmark}
We evaluate the performance of \MethodName{} on three complex and realistic benchmarks. Here, we provide a brief introduction to these benchmarks.

\textbf{WebShop}. A comprehensive benchmark designed to evaluate the decision-making and language understanding capabilities of autonomous agents within a simulated e-commerce environment. It challenges agents to perform complex, multi-step tasks such as searching, navigating product pages, and selecting specific attributes (e.g., size, color, or style) to satisfy a given user requirement. Unlike static datasets, WebShop requires agents to interact with a dynamic interface, bridging the gap between linguistic instructions and functional web actions. It serves as a rigorous testing ground for assessing an agent’s ability to reason and manage high-dimensional action spaces in realistic online scenarios.

\textbf{ALFWorld}. A large-scale benchmark for developing embodied AI agents that bridge the gap between language reasoning and physical execution. It maps high-level directives from the ALFRED dataset \cite{shridhar2020alfred} into interactive, text-based environments using TextWorld. The benchmark features six categories of household tasks that require agents to perceive environment states, plan long-horizon trajectories, and generalize their knowledge across diverse indoor scenes.

\textbf{AppWorld}. AppWorld serves as a rigorous testing ground for LLM-based agents, providing a controllable yet realistic environment for tool-use and task planning. The benchmark comprises a rich suite of "apps" with a unified authentication system and a persistent database, simulating a multi-user world. It challenges agents with tasks that require deep API composition, long-horizon planning, and the ability to handle interleaved dependencies between different services. By utilizing an execution-based evaluation metric, where success is determined by the actual side effects on the environment rather than mere text similarity, AppWorld addresses the limitations of static evaluation and provides a more accurate assessment of an agent’s reliability in real-world scenarios.

\subsection{Comparing Baselines}
We select nine baselines for comparison, including off-the-shelf models, prompt-based methods, and RL training approaches. Below, we provide a brief introduction to these baselines.

\textbf{Qwen3.5-397B}. A large-scale open-weights model with 397 billion parameters, known for its robust general capabilities and reasoning \cite{yang2025qwen3}.

\textbf{DeepSeek-V3.2}. Another open-weights Mixture-of-Experts model optimized for efficiency, coding, and mathematical reasoning \cite{liu2025deepseek}.

\textbf{Gemini-2.5-Pro}. A proprietary multimodal model featuring advanced long-context understanding and cross-modal reasoning \cite{comanici2025gemini}.

\textbf{ReAct}. A framework that combines reasoning traces and task-specific actions to enable interleaved decision-making \cite{yao2022react}.

\textbf{Reflexion}. An agentic architecture that utilizes linguistic self-reflection and trial-and-error to iteratively improve task performance \cite{shinn2023reflexion}.

\textbf{PPO}. The standard policy gradient algorithm used for stable and efficient reinforcement learning from human feedback \cite{schulman2017proximal}.

\textbf{RLOO}. An online RL method that leverages leave-one-out variance reduction to improve the efficiency of preference alignment \cite{ahmadian2024back}.

\textbf{GRPO} A critic-free reinforcement learning approach that optimizes policies using relative rewards within sampled groups \cite{shao2024deepseekmath}.

\textbf{GiGPO}. A reinforcement learning algorithm that achieves fine-grained multi-step credit assignment for LLM agents by estimating relative advantages at both the episode-level and the step-level via retroactively grouped identical environment states \cite{fenggroup}.

\subsection{Training Details}
\label{sec:details}

\textbf{Hyperparameters for ALFWorld.} We apply consistent hyperparameters across all methods: the maximum lengths for prompts and responses are capped at 2048 and 512 tokens, respectively, with a maximum of 50 environment interaction steps per episode. The learning rates for the Actor and PPO critic are set to $1 \times 10^{-6}$ and $1 \times 10^{-5}$. We employ a rule-based reward system where success receives a reward of 10, failure receives 0, and a penalty of -0.1 is applied to invalid actions. During the rollout process, PPO utilizes 128 independent environments, whereas the group reinforcement learning methods perform sampling across 16 groups of 8 environments each (totaling 128 environments). The rollout and validation temperatures are set to 1.0 and 0.4, respectively. Other parameters include a mini-batch size of 256 and a KL divergence coefficient of 0.01. For \MethodName{}, we set $\gamma = 0.95$ and fix $\omega = 1$.

\textbf{Hyperparameters for WebShop.} Identical hyperparameters are applied across all evaluated methods, with the maximum prompt and response lengths restricted to 4096 and 512 tokens, respectively. Each episode comprises a maximum of 15 environment steps. We set the learning rate at $1 \times 10^{-6}$ for the actor and $1 \times 10^{-5}$ for the critic (exclusive to PPO). A rule-based reward system is employed, where successes yield 10, failures yield 0, and invalid actions are penalized with -0.1. Following the ALFWorld setup, group-based RL methods utilize 128 environments in total (16 groups of 8 per rollout), whereas PPO employs 128 distinct environments. Temperatures for rollout and validation are maintained at 1.0 and 0.4. Furthermore, we use a mini-batch size of 64 and a KL-divergence loss coefficient of 0.01. For \MethodName{}, the weighting coefficient $\omega$ is fixed at 1 without further tuning, alongside a discount factor $\gamma$ of 0.95.

\textbf{Hyperparameters for AppWorld.} Identical hyperparameters are applied across all evaluated methods, with the maximum prompt and response lengths restricted to 13000 and 512 tokens, respectively. Each episode comprises a maximum of 30 environment steps. We set the learning rate at $1 \times 10^{-6}$ for the actor and $1 \times 10^{-5}$ for the critic (exclusive to PPO). A rule-based reward system is employed, where successes yield 10, failures yield 0, and invalid actions are penalized with -0.1. Following the ALFWorld setup, group-based RL methods utilize 128 environments in total (16 groups of 8 per rollout), whereas PPO employs 128 distinct environments. Temperatures for rollout and validation are maintained at 1.0 and 0.4. Furthermore, we use a mini-batch size of 64 and a KL-divergence loss coefficient of 0.01. For \MethodName{}, the weighting coefficient $\omega$ is fixed at 1 without further tuning, alongside a discount factor $\gamma$ of 0.95.

\textbf{Computing Details.} For ALFWorld and WebShop, We run Qwen2.5-1.5/7B experiments on 4×H100 GPUs, each for 100 iterations. For AppWorld, Qwen2.5-14B uses 8×H100 GPUs, for 50 iterations.

\subsection{Prompts}
Here we provide the prompts to guide reasoning and action of agents in RL on three benchmarks: WebShop (Figure \ref{fig:webshop_prompt}), ALFWorld (Figure \ref{fig:alfworld_prompt}), and AppWorld (Figure \ref{fig:appworld_prompt}).

\begin{figure}[h]
\centering
\resizebox{0.85\textwidth}{!}{
\begin{tcolorbox}[colback=gray!5!white, colframe=black!75!black, 
title=Prompt Template for ALFWorld, boxrule=0.3mm, width=\textwidth, arc=3mm, auto outer arc=true]
You are an expert agent operating in the ALFRED embodied environment. Your task is to: \textcolor{brown}{\{task\_description\}}. Prior to this step, you have already taken \textcolor{brown}{\{step\_count\}} step(s). Below are the most recent \textcolor{brown}{\{history\_length\}} observations and the corresponding actions you took: \textcolor{brown}{\{action\_history\}}. You are now at step \textcolor{brown}{\{current\_step\}} and your current observation is: \textcolor{brown}{\{current\_observation\}}. Your admissible actions of the current situation are: [\textcolor{brown}{\{admissible\_actions\}}].

Now it's your turn to take an action.
You should first reason step-by-step about the current situation. This reasoning process MUST be enclosed within \textcolor{deepgreen}{<think>} \textcolor{deepgreen}{</think>} tags.
Once you've finished your reasoning, you should choose an admissible action for current step and present it within \textcolor{deeppurple}{<action>} \textcolor{deeppurple}{</action>} tags.
\end{tcolorbox}
}
\caption{Prompt template for ALFWorld.}
\label{fig:alfworld_prompt}
\end{figure}

\begin{figure}[h]
\centering
\resizebox{0.85\textwidth}{!}{
\begin{tcolorbox}[colback=gray!5!white, colframe=black!75!black, 
title=Prompt Template for WebShop, boxrule=0.3mm, width=\textwidth, arc=3mm, auto outer arc=true]
You are an expert autonomous agent operating in the WebShop e‑commerce environment. Your task is to: \textcolor{brown}{\{task\_description\}}. Prior to this step, you have already taken \textcolor{brown}{\{step\_count\}} step(s). Below are the most recent \textcolor{brown}{\{history\_length\}} observations and the corresponding actions you took: \textcolor{brown}{\{action\_history\}}. You are now at step \textcolor{brown}{\{current\_step\}} and your current observation is: \textcolor{brown}{\{current\_observation\}}. Your admissible actions for the current situation are: [\textcolor{brown}{\{available\_actions\}}].

Now it's your turn to take one action for the current step.
You should first reason step-by-step about the current situation, then think carefully which admissible action best advances the shopping goal. This reasoning process MUST be enclosed within \textcolor{deepgreen}{<think>} \textcolor{deepgreen}{</think>} tags.
Once you've finished your reasoning, you should choose an admissible action for current step and present it within \textcolor{deeppurple}{<action>} \textcolor{deeppurple}{</action>} tags.
\end{tcolorbox}
}
\caption{Prompt template for WebShop.}
\label{fig:webshop_prompt}
\end{figure}

\begin{figure}[h]
\centering
\resizebox{0.85\textwidth}{!}{
\begin{tcolorbox}[colback=gray!5!white, colframe=black!75!black, 
title=Prompt Template for AppWorld, boxrule=0.3mm, width=\textwidth, arc=3mm, auto outer arc=true]
I am your supervisor and you are a super intelligent AI Assistant whose job is to achieve my day-to-day tasks completely autonomously.

To do this, you will need to interact with app/s (e.g., spotify, venmo, etc) using their associated APIs on my behalf. For this you will undertake a \textit{multi-step conversation} using a python REPL environment. That is, you will write the python code and the environment will execute it and show you the result, based on which, you will write python code for the next step and so on, until you've achieved the goal. This environment will let you interact with app/s using their associated APIs on my behalf.

Here are three key APIs that you need to know to get more information

\# To get a list of apps that are available to you.

print(apis.api\_docs.show\_app\_descriptions())

\# To get the list of apis under any app listed above, e.g. supervisor

print(apis.api\_docs.show\_api\_descriptions(app\_name='supervisor'))

\# To get the specification of a particular api, e.g. supervisor app's show\_account\_passwords

print(apis.api\_docs.show\_api\_doc(app\_name='supervisor', api\_name='show\_account\_passwords'))

The code will be executed in an interactive shell. Each code execution will produce an output that you can use in subsequent calls. Using these APIs, you can now generate code, that the environment will execute, to solve the task.

-----------------------------

Here is an example:

...

-----------------------------

Key Instructions and Disclaimers:

1. The email addresses, access tokens and variables (e.g. spotify\_password) in the example above were only for demonstration. Obtain the correct information by calling relevant APIs yourself.

2. Only generate valid code blocks, i.e., do not put them in ```...``` or add any extra formatting. Any thoughts should be put as code comments.

3. You can use the variables from the previous code blocks in the subsequent code blocks.

...

16. Once you believe the task is complete, you MUST call `apis.supervisor.complete\_task()` to finalize it. If the task requires an answer, provide it using the answer argument — for example, `apis.supervisor.complete\_task(answer=<answer>)`. For tasks that do not require an answer, either omit the argument. The task will not end automatically — it will remain open until you explicitly make this call.

Using these APIs, now begin writing code cells step-by-step to solve the actual task:

My name is: \textcolor{brown}{\{supervisor\_first\_name\}} \textcolor{brown}{\{supervisor\_last\_name\}}. My personal email is \textcolor{brown}{\{supervisor\_email\}} and phone number is \textcolor{brown}{\{supervisor\_phone\_number\}}.

Your task is: \textcolor{brown}{\{task\_description\}}

Prior to this step, you have already taken \textcolor{brown}{\{step\_count\}} step(s). Below are the most recent \textcolor{brown}{\{history\_length\}} codes you generated and the corresponding environment return: 
\textcolor{brown}{\{action\_history\}}

Now you are at step \textcolor{brown}{\{current\_step\}} and it's your turn to generate code for this step.
First, you MUST carefully reflect on the history of interactions and the most recent error messages. Then, reason about what should be done next, which APIs to call, what arguments to use, and how to build your code block to complete the task. This reasoning process MUST be enclosed within \textcolor{green}{<think>} \textcolor{green}{</think>} tags.
Once you've finished your reflexion and reasoning, you present the solution code body within \textcolor{purple}{<code>} \textcolor{purple}{</code>} tags.
\end{tcolorbox}
}
\caption{Prompt template for AppWorld.}
\label{fig:appworld_prompt}
\end{figure}

\section{Case Study}
\label{sec:case}
\begin{figure}[!t]
\centering
\includegraphics[width=1.0\columnwidth]{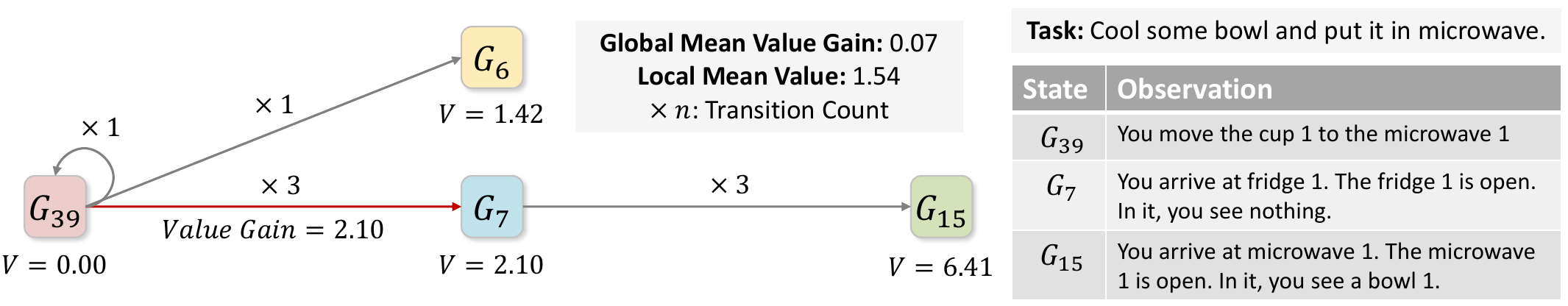}
\caption{A case study illustrating the superiority of the edge-centric advantage over local group comparisons. While local group comparisons provide the relative advantage between different actions within the same state, they fail to evaluate the absolute value improvement brought by an action from a global graph perspective.}
\label{fig:case}
\vspace{-0.5cm}
\end{figure}

To demonstrate the superiority of edge-centric advantage ($A^{EC}$) over local group comparisons, we provide a case graph constructed during the advantage estimation phase of \MethodName{} training on ALFWorld. As shown in Figure \ref{fig:case}, to estimate the advantage of transitioning from state node 39 to node 7, the local group comparison method calculates the average value of all possible destination nodes as a baseline and measures the relative advantage of node 7's value against this average. While this provides a fine-grained credit assignment, it entirely ignores the intrinsic value of the source node and the actual value increment driven by the state transition. Consequently, this outcome-focused approach fails to evaluate the absolute contribution of an action. In contrast, $A^{EC}$ evaluates an action's contribution to the overall goal by comparing the absolute value increments across all actions within the global graph, thereby overcoming the limited scope and destination-node dependency of local comparisons. 

Specifically, the task is to \textit{cool some bowl and put it in microwave}. The observation at node 39 is \textit{You move the cup 1 to the microwave 1}, which completely deviates from the task objective. Conversely, the observation at node 7 is \textit{You arrive at fridge 1. The fridge 1 is open. In it, you see nothing}, proving the agent has located the correct tool to \textit{cool bowl}. This represents a massive leap forward toward the goal and thus merits a large positive advantage. However, under local group comparison, node 7 receives a marginal positive advantage of only 0.55 because its value is simply close to the average of all possible destination nodes. In contrast, $A^{EC}$ accounts for the low value of the source node and standardizes it against the average value increments of all actions, assigning a substantial positive advantage of 2.1.

\clearpage
\section{State Grouping under Observation Noise}
\label{sec:noisy_grouping}

\textbf{Experimental setup.} To test whether exact matching is essential, we construct a noisy ALFWorld variant by inserting task-irrelevant sentences, such as descriptions of outside noise or electromagnetic interference, into every returned observation. We train Qwen2.5-1.5B on this variant, merging observations when their Python \texttt{SequenceMatcher} similarity exceeds 0.95. We compare this setting with exact matching on the original observations without noise.

\textbf{Results.} Table~\ref{tab:noisy_grouping} reports the success rates for both settings. Similarity matching under observation noise achieves an overall success rate of 95.3\%, compared with 95.0\% for exact matching without noise. Although individual task categories vary, overall performance is preserved. This result supports the flexibility of \MethodName{} beyond byte-identical observations when nuisance text is present.

\begin{table}[!htbp]
\centering
\caption{Success rates (\%) on ALFWorld under different observation and grouping settings.}
\label{tab:noisy_grouping}
\small
\setlength{\tabcolsep}{4pt}
\begin{tabular}{lrrrrrrr}
\toprule
Setting & Pick & Look & Clean & Heat & Cool & Pick2 & All \\
\midrule
No noise + exact matching & 97.2 & 96.3 & 95.1 & 94.9 & 97.1 & 91.4 & 95.0 \\
Noise + similarity matching & 94.0 & 81.8 & 100.0 & 100.0 & 100.0 & 88.9 & 95.3 \\
\bottomrule
\end{tabular}
\end{table}

\section{Empirical Comparison with GEPO}
\label{sec:gepo_comparison}

GEPO~\cite{yuan2025graph} uses $N=16$ rollouts per group in its main experiments, whereas all main-paper \MethodName{} results use $N=8$. We therefore additionally train \MethodName{} with $n=16$ on ALFWorld using Qwen2.5-1.5B, and compare the $n=8$ setting on WebShop using Qwen2.5-7B, for which GEPO also reports results. The GEPO results below are taken from its reported experiments.

\begin{table}[!htbp]
\centering
\caption{Success rates (\%) on ALFWorld, comparing GEPO and \MethodName{}.}
\label{tab:gepo_alfworld}
\small
\begin{tabular}{lrrrrrrr}
\toprule
Method & Pick & Look & Clean & Heat & Cool & Pick2 & All \\
\midrule
GEPO ($N=16$) & 99.2 & 84.5 & 96.8 & 88.5 & 83.6 & 92.8 & 92.9 \\
\MethodName{} ($N=8$) & 97.2 & 96.3 & 95.1 & 94.9 & 97.1 & 91.4 & 95.0 \\
\MethodName{} ($N=16$) & 100.0 & 100.0 & 100.0 & 100.0 & 89.5 & 95.2 & 97.7 \\
\bottomrule
\end{tabular}
\end{table}

As shown in Table~\ref{tab:gepo_alfworld}, \MethodName{} exceeds GEPO on ALFWorld by 2.1 percentage points while using half as many rollouts per group (95.0\% vs.\ 92.9\%). At the matched rollout count of $N=16$, \MethodName{} achieves 97.7\%, a gain of 4.8 points over GEPO.

\begin{table}[!htbp]
\centering
\caption{Success rates (\%) on WebShop with Qwen2.5-7B.}
\label{tab:gepo_webshop}
\begin{tabular}{lr}
\toprule
Method & Success rate \\
\midrule
GEPO ($N=8$) & 78.5 \\
\MethodName{} ($N=8$) & 78.3 \\
\bottomrule
\end{tabular}
\end{table}

On WebShop, the two methods achieve comparable success rates at $N=8$, differing by only 0.2 percentage points (Table~\ref{tab:gepo_webshop}). GEPO reports additional training overhead of approximately 15\% at $N=8$ and 35\% at $N=16$, while the overhead measured for \MethodName{} in our main-paper setting is approximately 0.4\% per training step (Figure~\ref{fig:compute}). These results show comparable or higher success rates for \MethodName{} across the evaluated settings, with substantially lower reported computational overhead.

\clearpage
\section{Diagnostic Validation of the Edge-Centric Advantage}
\label{sec:edge_diagnostics}

We use ALFWorld and Qwen2.5-1.5B to examine the edge-centric advantage from two perspectives: the credit assigned to actions on a shortest path to success, and its correlation with action importance.

\subsection{Advantages of Critical and Non-Critical Actions}

For each constructed task graph, we find a shortest path from the initial state to a success state, label actions on this path as critical, and label the remaining actions as non-critical. For this diagnostic, we report the unnormalized edge-centric advantage, denoted $A^{EC}_{\mathrm{raw}}$, before the global standardization in Eq.~\ref{eq:aec}. We compare its average over critical and non-critical actions within each task.

\begin{table}[!htbp]
\centering
\caption{Unnormalized edge-centric advantages for critical and non-critical actions on ALFWorld.}
\label{tab:critical_actions}
\begin{tabular}{lr}
\toprule
Metric & Value \\
\midrule
Tasks where mean critical $A^{EC}_{\mathrm{raw}}$ exceeds mean non-critical $A^{EC}_{\mathrm{raw}}$ & 91.0\% \\
Average critical $A^{EC}_{\mathrm{raw}}$ & $+0.7049$ \\
Average non-critical $A^{EC}_{\mathrm{raw}}$ & $+0.2394$ \\
Average difference (critical $-$ non-critical) & $+0.4655$ \\
\bottomrule
\end{tabular}
\end{table}

As shown in Table~\ref{tab:critical_actions}, critical actions receive higher average advantages in 91.0\% of tasks. Their average advantage is nearly three times that of non-critical actions, with a mean difference of $+0.4655$. This diagnostic supports the ability of edge-centric credit assignment to prioritize actions on short paths to task completion.

\subsection{Correlation with Action Importance}

We define an action's importance as the task success rate following its execution. We calculate the episode-level, node-centric, and edge-centric advantages and measure their Pearson correlation $r$ and Spearman rank correlation $\rho$ with action importance.

\begin{table}[!htbp]
\centering
\caption{Correlation between advantage estimates and action importance on ALFWorld.}
\label{tab:action_importance}
\begin{tabular}{lrr}
\toprule
Advantage method & Pearson $r$ & Spearman $\rho$ \\
\midrule
Episode-level & $+0.0304$ & $+0.0238$ \\
Node-centric & $+0.0492$ & $+0.0625$ \\
Edge-centric & $+0.2531$ & $+0.1636$ \\
\bottomrule
\end{tabular}
\end{table}

Table~\ref{tab:action_importance} shows that the edge-centric signal has approximately $5.1\times$ and $8.3\times$ the Pearson correlation of the node-centric and episode-level signals, respectively, and $2.6\times$ and $6.9\times$ their Spearman correlation. Although the absolute correlations are modest, the stronger association supports the ability of the edge-centric signal to identify actions associated with higher task success rates.


\end{document}